\documentclass[journal]{IEEEtran}
\usepackage{cite}
\usepackage{amsmath,amsfonts}
\usepackage{hyperref}
\usepackage{algorithmic}
\usepackage{algorithm}
\usepackage{array}
\usepackage[caption=false,font=normalsize,labelfont=sf,textfont=sf]{subfig}
\usepackage{textcomp}
\usepackage{stfloats}
\usepackage{url}
\usepackage{verbatim}
\usepackage{graphicx}
\usepackage{cite}
\usepackage{booktabs}
\usepackage{multirow}
\usepackage{bm}
\usepackage{amssymb}
\usepackage[edges]{forest}
\usepackage{bbding}
\usepackage{soul}
\usepackage[T1]{fontenc}
\usepackage[utf8]{inputenc}
\usepackage[english]{babel}
\usepackage{pifont}       
\usepackage{bbding}       
\usepackage{xcolor}

\begin{document}
\title{
\includegraphics[height=1.5em]{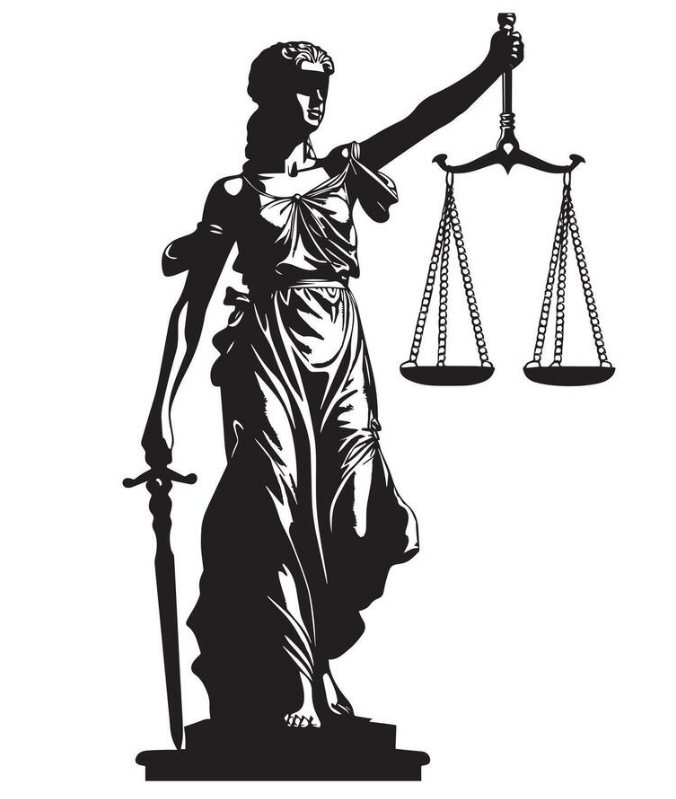}The Scales of Justitia: A Comprehensive Survey on Safety Evaluation of LLMs
}

\author{Songyang~Liu,
        Chaozhuo~Li$^*$,
        Jiameng~Qiu, 
        Xi~Zhang, 
        Feiran~Huang,
        Litian~Zhang,
        Yiming~Hei,
        and~Philip~S.~Yu
\thanks{S. Liu, C. Li, X. Zhang are with School of Cyberspace Security, Beijing University of Posts and Telecommunications, Beijing, China (Email: liusyang@bupt.edu.cn; lichaozhuo@bupt.edu.cn; zhangx@bupt.edu.cn).}
\thanks{J. Qiu, F. Huang are with School of Cyberspace Security, Jinan University, Guangzhou, China (Email: qiujiameng@stu2024.jnu.edu.cn, huangfr@jnu.edu.cn).}
\thanks{L. Zhang is with School of Cyberspace Security, Beihang University, Beijing, China (Email: litianzhang@buaa.edu.cn).}
\thanks{Y. Hei is with China Academy of Information and Communications Technology, Beijing, China (Email: heiyiming@caict.ac.cn)}
\thanks{P.S. Yu is with Department of Computer Science, University of Illinons at Chicago, Chicago, USA (Email: psyu@uic.edu).}
\thanks{Chaozhuo Li is the Corresponding Author.}
}

\markboth{Journal of \LaTeX\ Class Files,~Vol.~14, No.~8, August~2015}%
{Shell \MakeLowercase{\textit{et al.}}: Bare Demo of IEEEtran.cls for IEEE Journals}


\maketitle

\begin{abstract}
With the rapid advancement of artificial intelligence, Large Language Models (LLMs) have shown remarkable capabilities in Natural Language Processing (NLP), including content generation, human-computer interaction, machine translation, and code generation. However, their widespread deployment has also raised significant safety concerns. In particular, LLM-generated content can exhibit unsafe behaviors such as toxicity, bias, or misinformation, especially in adversarial contexts, which has attracted increasing attention from both academia and industry. Although numerous studies have attempted to evaluate these risks, a comprehensive and systematic survey on safety evaluation of LLMs is still lacking. This work aims to fill this gap by presenting a structured overview of recent advances in safety evaluation of LLMs. Specifically, we propose a four-dimensional taxonomy: (i) \textit{Why to evaluate}, which explores the background of safety evaluation of LLMs, how they differ from general LLMs evaluation, and the significance of such evaluation; (ii) \textit{What to evaluate}, which examines and categorizes existing safety evaluation tasks based on key capabilities, including dimensions such as toxicity, robustness, ethics, bias and fairness, truthfulness, and related aspects; (iii) \textit{Where to evaluate}, which summarizes the evaluation metrics, datasets and benchmarks currently used in safety evaluations; (iv) \textit{How to evaluate}, which reviews existing mainstream evaluation methods based on the roles of the evaluators and some evaluation frameworks that integrate the entire evaluation pipeline. Finally, we identify the challenges in safety evaluation of LLMs and propose promising research directions to promote further advancement in this field. We emphasize the necessity of prioritizing safety evaluation to ensure the reliable and responsible deployment of LLMs in real-world applications.
\end{abstract}


%
\IEEEpeerreviewmaketitle

\section{Introduction}
\IEEEPARstart{A}{s} artificial intelligence continues to advance at an unprecedented pace, Large Language Models (LLMs) have rapidly emerged as a focal point in both academia and industry. These models have achieved significant breakthroughs in Natural Language Processing (NLP) and demonstrated strong capabilities across diverse applications, including content generation \cite{chang2024survey}, human-computer interaction \cite{kotian2024systematic}, machine translation \cite{zhao2024handcrafted}, and code generation \cite{chen2024survey}, among others \cite{yan2025beyond, yin2024unleash}. By leveraging massive training corpora, LLMs can generate coherent, human-like text while exhibiting a degree of comprehension and reasoning ability \cite{mondorf2024beyond}, thus driving rapid progress in the broader field of artificial intelligence.

\begin{figure}[t!]
    \centering
    \includegraphics[width=\linewidth]{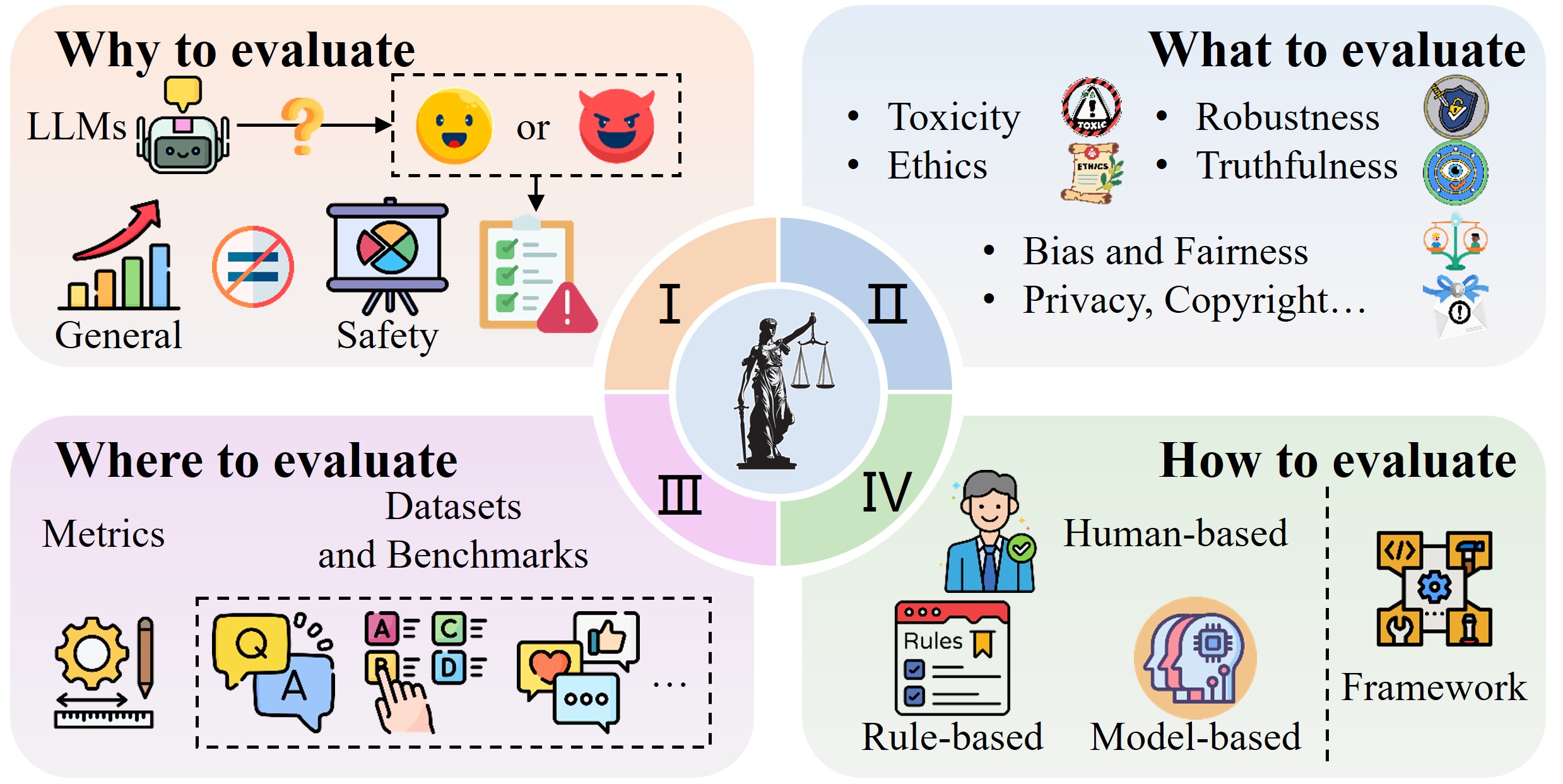}
    \caption{Overview of the Comprehensive and Systematic Survey on Safety Evaluation of LLMs.}
    \label{fig:overview}
\end{figure}

However, like any emerging technology, the widespread application of LLMs has introduced critical safety challenges. Recent studies have demonstrated that LLMs are prone to generating harmful content, such as toxicity\cite{zou2023universal}, bias\cite{zhang2023chatgpt} and false information \cite{zhou2024finefake}, with such issues often exacerbated under adversarial scenarios. In addition, attack methods targeting LLMs safety, such as prompt injection attacks and jailbreak attacks \cite{pu2024baitattack, pu2024feint}, have become increasingly sophisticated, seeking to bypass alignment mechanisms and induce unsafe behavior from LLMs \cite{liu2024exploring}. These vulnerabilities not only degrade user experience but also raise profound ethical and legal risks, offering opportunities for malicious exploitation. Consequently, ensuring the safety of LLMs through systematic evaluation has become an urgent priority.

\begin{table*}[!ht]
\centering
\caption{A Comparative Analysis of Related Surveys and Studies on Safety Evaluations of LLMs}
\label{tab:all_works1}
\setlength\tabcolsep{6.8pt}
\renewcommand{\arraystretch}{1.2} 
\resizebox{\linewidth}{!}{
\begin{tabular}{lc|cccccccc}
\toprule
\multirow{2}[2]{*}{\textbf{Papers}} & \multirow{2}[2]{*}{\textbf{Year}} & \multicolumn{6}{c}{\textbf{Key Objectives of Safety Evaluation}} \\ 
\cmidrule(lr){3-8}
& & Taxonomy of Tasks & Metrics & Datasets and Benchmarks & Taxonomy of Evaluators & Frameworks & Challenges and Future Directions\\ 
\midrule
Guo \textit{et al.} \cite{guo2023evaluating} & 2023 & \textcolor{green}{\ding{51}} & \textcolor{red}{\ding{55}} & \textcolor{green}{\ding{51}} & \textcolor{red}{\ding{55}} & \textcolor{red}{\ding{55}} & \textcolor{red}{\ding{55}}\\

Dong \textit{et al.} \cite{dong2024attacks} & 2024 & \textcolor{red}{\ding{55}} & \textcolor{green}{\ding{51}} & \textcolor{green}{\ding{51}} & \textcolor{red}{\ding{55}} & \textcolor{red}{\ding{55}} & \textcolor{green}{\ding{51}}\\

Yi \textit{et al.} \cite{yi2024jailbreak} & 2024 & \textcolor{red}{\ding{55}} & \textcolor{green}{\ding{51}} & \textcolor{green}{\ding{51}} & \textcolor{red}{\ding{55}} & \textcolor{green}{\ding{51}} & \textcolor{red}{\ding{55}}\\

Ran \textit{et al.} \cite{ran2024jailbreakeval} & 2024 & \textcolor{red}{\ding{55}} & \textcolor{red}{\ding{55}} & \textcolor{red}{\ding{55}} & \textcolor{green}{\ding{51}} & \textcolor{red}{\ding{55}} & \textcolor{red}{\ding{55}}\\

R{\"o}ttger \textit{et al.} \cite{rottger2025safetyprompts} & 2025 & \textcolor{red}{\ding{55}} & \textcolor{red}{\ding{55}} & \textcolor{green}{\ding{51}} & \textcolor{red}{\ding{55}} & \textcolor{red}{\ding{55}} & \textcolor{red}{\ding{55}}\\

\textbf{Ours} & 2025 & \textcolor{green}{\ding{51}} & \textcolor{green}{\ding{51}} & \textcolor{green}{\ding{51}} & \textcolor{green}{\ding{51}} & \textcolor{green}{\ding{51}} & \textcolor{green}{\ding{51}}\\

\bottomrule
\end{tabular}
}
\end{table*}

Safety evaluation of LLMs is the systematic process of examining whether a model can resist harmful instructions, avoid generating unsafe outputs, and maintain robustness under both benign and adversarial interactions, thereby assessing the degree of risk that LLMs may pose to social order, public interest, and human well-being. The emergence of this research area stems from a growing recognition and urgent need to address the potential risks posed by these models and their societal impacts. Within the broader lifecycle of LLMs research, development, and deployment, safety evaluation plays a crucial role. Unlike general ability evaluation of LLMs \cite{guo2023evaluating,chang2024survey,peng2024survey}, which focus primarily on general capabilities such as language understanding, generation, and reasoning, safety evaluation places particular emphasis on the risks and ethical concerns associated with model outputs. The evaluation covers a range of critical dimensions, including toxicity, robustness, morality, bias and fairness, and credibility. By conducting a systematic safety evaluation of LLMs, potential risks can be identified and mitigated in a timely manner, thereby ensuring the safety of these models in practical applications. This rigorous evaluation process is essential for enhancing user trust and preventing societal issues that may arise from model deficiencies, ultimately fostering the healthy development and widespread adoption of LLMs technologies.

While numerous researchers have actively explored the safety evaluation of LLMs in recent years, the field still lacks a comprehensive and systematic review. As shown in Table \ref{tab:all_works1}, most existing relevant surveys concentrate on general ability evaluation of LLMs \cite{guo2023evaluating} on specific topics such as jailbreak attacks \cite{dong2024attacks, yi2024jailbreak}, with only limited coverage of safety evaluation, failing to form a systematic classification framework and content synthesis. Furthermore, existing studies often remain fragmented, addressing single aspects such as datasets and benchmarks \cite{rottger2025safetyprompts} or evaluators \cite{ran2024jailbreakeval}, lacking a holistic analysis of the entire pipeline, which includes evaluation methods, technical challenges, and future directions. This fragmentation underscores the urgent need for a structured and integrative survey on safety evaluation of LLMs.

Against this backdrop, this survey aims to provide a comprehensive and systematic review of recent advancements in safety evaluation of LLMs, thereby addressing this research gap. Specifically, as shown in Figure~\ref{fig:overview}, we delve into existing work from the following four dimensions: (i) \textit{Why to Evaluate}, which elucidates the background of safety evaluation, distinguishing it from general ability evaluation. This section underscores the significance of safety evaluations in ensuring that LLMs can be responsibly deployed in real-world applications, highlighting the potential risks associated with unassessed models; (ii) \textit{What to Evaluate}, which summarizes the primary tasks associated with safety evaluation of LLMs, presenting a detailed classification that encompasses various facets such as toxicity, robustness, ethics, bias and fairness, truthfulness, and more. (iii) \textit{Where to Evaluate}, which compiles current commonly used evaluation metrics, and categorizes the datasets and benchmarks employed in the field. The aim of this section is to provide researchers with a comprehensive reference for selecting appropriate evaluation criteria, thereby facilitating faster progress in safety evaluation research; and (iv) \textit{How to Evaluate}, categorizes evaluation methods based on the roles of the evaluators, whether they are automated or manual. It also reviews several evaluation frameworks, providing insights into the methodologies and best practices for conducting safety evaluations. Finally, we highlight the key challenges in this field and outline promising avenues for future research.

The contributions of this paper can be summarized as follows:
\begin{itemize}
    \item \textbf{Comprehensive Review:} To the best of our knowledge, this is the \textbf{first} survey to present a systematic and holistic synthesis of recent advancements in safety evaluation of LLMs, addressing a significant gap in the literature.
    \item \textbf{Clear Classification Framework:} We establish a detailed taxonomy that delineates the primary tasks of safety evaluation across multiple dimensions, providing conceptual clarity for the research community.
    \item \textbf{Consolidated Evaluation Resources:} We compile and categorize existing evaluation metrics, datasets, benchmarks, and methodologies, offering a consolidated reference for future research and practice.
    \item \textbf{Future Research Directions:} We highlight key challenges currently facing the field and propose potential research directions to promote the responsible and sustainable development of safety evaluation.
\end{itemize}

\newcolumntype{P}[1]{>{\RaggedRight\arraybackslash}p{#1}}

\definecolor{root-color}{HTML}{F3CDF4}
\definecolor{child-one-color}{HTML}{FBF7E7}
\definecolor{child-two-color}{HTML}{E3F3DF}
\definecolor{child-three-color}{HTML}{E2ECF7}
\definecolor{child-four-color}{HTML}{FDE7D0}
\definecolor{child-five-color}{HTML}{F6DADC}

\definecolor{root-line-color}{HTML}{FFCCCC}
\definecolor{child-one-line-color}{HTML}{E9D070}
\definecolor{child-two-line-color}{HTML}{64BA7D}
\definecolor{child-three-line-color}{HTML}{79ADD3}
\definecolor{child-four-line-color}{HTML}{EE9061}
\definecolor{child-five-line-color}{HTML}{E99FA4}

\definecolor{edge-color}{HTML}{000000}


\definecolor{hidden-draw}{RGB}{20,68,106}
\definecolor{hidden-pink}{RGB}{255,245,247}
\definecolor{red}{RGB}{255,0,0}


\definecolor{hidden-draw}{RGB}{0,0,0}
\definecolor{hidden-pink}{RGB}{255,182,193}






\newcommand\vr[1]{\todo[author=VR,color=orange!40]{#1}}
\newcommand\vril[1]{\todo[author=VR,color=orange!40,inline]{#1}}

\begin{figure*}[ht!]
	\centering
	\resizebox{\textwidth}{!}{
		\begin{forest}
			for tree={
                    grow=east,
                    reversed=true,
                    anchor=base west,
                    parent anchor=east,
                    child anchor=west,
                    base=center,
                    font=\large,
                    rectangle,
                    rounded corners,
                    align=center,
                    text centered,
                    minimum width=5em,
                    edge+={edge-color, line width=1pt},
                    s sep=3pt,
                    inner xsep=3pt,
                    inner ysep=3pt,
                    line width=1pt,
                    edge path={
                        \noexpand\path [\forestoption{edge}]
                        (!u.parent anchor) -- +(.2cm,0) |- (.child anchor)
                        \forestoption{edge label};
                    },
                },
			where level=1{
				text width=13em,
				font=\normalsize,
			}{},
			where level=2{
				text width=20em,
				font=\normalsize,
			}{},
			where level=3{
				minimum width=28em,
				font=\normalsize,
			}{},
			[
				\textbf{LLMs Safety Evaluation},
				for tree={fill=root-color},
				[
					Why to evaluate \hyperlink{WHY TO EVALUATE}{$\S$ II},   
					for tree={fill=child-one-color},
					[
						Background and Foundations \hyperlink{Background and Foundations}{(A)},
						for tree={fill=child-one-color}
					]
					[
                            {LLMs Evaluation vs}\\
					    {LLMs Safety Evaluation}
						\hyperlink{LLMs Evaluation vs LLMs Safety Evaluation}{(B)},
						for tree={fill=child-one-color},
					]
                        [
						Importance \hyperlink{Importance}{(C)},
						for tree={fill=child-one-color},
					]
				]
				[
					What to evaluate \hyperlink{WHAT TO EVALUATE}{$\S$ III},  
					for tree={fill=child-two-color},
					[
						Toxicity \hyperlink{Toxicity}{(A)},
						for tree={fill=child-two-color},
						[
							ToxiGen \cite{hartvigsen2022toxigen};
							ToxiChat \cite{lin2023toxicchat}\\
                                LifeTox \cite{kim2023lifetox};
                                PolygloToxicityPrompts \cite{jain2024polyglotoxicityprompts}\\
                                Red-Eval \cite{bhardwaj2023red};
                                FFT \cite{cui2023fft}\\
                                zhuo \textit{et al.} \cite{zhuo2023red},
							for tree={fill=child-two-color}
						]
					]
					[
						Robustness \hyperlink{Robustness}{(B)},
						for tree={fill=child-two-color},
						[
							Chen \textit{et al.} \cite{chen2024characterizing};
                                DecodingTrust \cite{wang2023decodingtrust}\\
                                ASSERT \cite{mei2023assert};
                                JailbreakBench \cite{chao2024jailbreakbench}\\
                                HarmBench \cite{mazeika2024harmbench};
                                Röttger \textit{et al.} \cite{rottger2023xstest}\\
                                An \textit{et al.} \cite{an2024automatic};
                                Varshney \textit{et al.} \cite{varshney2023art},
							for tree={fill=child-two-color}
						]
					]
                        [
						Ethics \hyperlink{Ethics}{(C)},
						for tree={fill=child-two-color},
						[
							TechHazardQA \cite{banerjee2024ethical};
                                LocalValueBench \cite{meadows2024localvaluebench}\\
                                MoralBench \cite{ji2024moralbench};
                                CVALUES \cite{xu2023cvalues}\\
                                Sun \textit{et al.} \cite{sun2023safety};
                                Hendrycks \textit{et al.} \cite{hendrycks2020aligning},
							for tree={fill=child-two-color}
						]
					]
                        [
						Bias and Fairness \hyperlink{Bias and Fairness}{(D)},
						for tree={fill=child-two-color},
						[
							Cantini \textit{et al.} \cite{cantini2024large};
                                Zhang \textit{et al.} \cite{zhang2023chatgpt}\\
                                Wang \textit{et al.} \cite{wang2024large};
                                CEB \cite{wang2024ceb}\\
                                Huang \textit{et al.} \cite{huang2023cbbq};
                                Cui \textit{et al.} \cite{cui2023fft}\\
                                Azeem \textit{et al.} \cite{azeem2024llm};
                                Wang \textit{et al.} \cite{wang2023large}\\
                                Leidinger \textit{et al.} \cite{leidinger2024llms},
							for tree={fill=child-two-color}
						]
					]
                        [
						Truthfulness \hyperlink{Truthfulness}{(E)},
						for tree={fill=child-two-color},
						[
							AI-LIEDAR \cite{su2024ai};
                                THaMES \cite{liang2024thames}\\
                                ReEval \cite{yu2024reeval};
                                HaluEval \cite{li2023halueval}\\
                                Iqbal \textit{et al.} \cite{iqbal2024openfactcheck};
                                GraphEval \cite{sansford2024grapheval}\\
                                KG-FPQ \textit{et al.} \cite{zhu2024kg},
							for tree={fill=child-two-color}
						]
					]
                        [
						Other \hyperlink{Other}{(F)},
						for tree={fill=child-two-color},
						[
							LLM-PBE \cite{li2024llm};
                                MedSafetyBench \cite{medsafety}\\
                                CYBERSECEVAL \cite{bhatt2023purple};
                                CWEVAL \cite{peng2025cweval}\\
                                SafeGenBench \cite{li2025safegenbench};
                                SHIELD \cite{liu2024shield}\\
                                Trident-Bench \cite{hui2025trident},
							for tree={fill=child-two-color}
						]
					]
				]
				[
					Where to evaluate \hyperlink{WHERE TO EVALUATE}{$\S$ IV},  
					for tree={fill=child-three-color},
					[
						Metrics \hyperlink{Metrics}{(A)},
						for tree={fill=child-three-color},
						[
							Cai \textit{et al.} \cite{cai2024take};
                                StrongReject \cite{souly2024strongreject}\\
                                Chen \textit{et al.} \cite{chen2024characterizing};
                                Leidinger \textit{et al.} \cite{leidinger2024llms}\\
							Mei \textit{et al.} \cite{mei2024not},
							for tree={fill=child-three-color}
						]
					]
					[
						Datasets and Benchmarks \hyperlink{Datasets and Benchmarks}{(B)},
						for tree={fill=child-three-color},
						[
							Chu \textit{et al.} \cite{chu2024comprehensive};
                                R-Judge \cite{yuan2024r}\\
                                JailbreakBench \cite{chao2024jailbreakbench};
                                StrongREJECT \cite{souly2024strongreject}\\
                                HarmBench \cite{mazeika2024harmbench};
                                Do-Not-Answer \cite{wang2024not}\\
                                AdvBench \cite{zou2023universal};
                                CoSafe \cite{yu2024cosafe}\\
                                Chen \textit{et al.} \cite{chen2024characterizing};
                                TechHazardQA \cite{banerjee2024ethical}\\
                                FFT \cite{cui2023fft};
                                JailJudge ID \cite{liu2024jailjudge}\\
                                SORRY-Bench \cite{xie2024sorry};
                                HarmfulQA \cite{bhardwaj2023red}\\
                                AegisSafetyDataset \cite{ghosh2024aegis};
                                SALAD-Bench \cite{li2024salad}\\
                                ALERT \cite{tedeschi2024alert};
                                WildGuardMix \cite{han2024wildguard}\\
                                Do Anything Now \cite{shen2023anything};
                                CHiSafetyBench \cite{zhang2024chisafetybench}\\
                                Wang \textit{et al.} \cite{wang2024chinese};
                                CValues \cite{xu2023cvalues}\\
                                JailBench \cite{liu2025jailbench};
                                CRiskEval \cite{shi2024criskeval}\\
                                Safety Prompts \cite{sun2023safety};
                                ChineseSafe \cite{zhang2024chinesesafe}\\
                                Latent Jailbreak \cite{qiu2023latent};
                                SafetyBench \cite{zhang2024safetybench}\\
                                S-Eval \cite{yuan2024s};
                                AraTrust \cite{alghamdi2024aratrust}\\
                                AnswerCarefully \cite{suzuki2025answercarefully};
                                XSafety \cite{wang2024all}\\
                                JailJudge OOD \cite{liu2024jailjudge};
                                PolyglToxicityPrompts \cite{jain2024polyglotoxicityprompts},
							for tree={fill=child-three-color}
						]
					]
				]
				[
					How to evaluate \hyperlink{HOW TO EVALUATE}{$\S$ V}, 
					for tree={fill=child-four-color},
					[
						Human-based Evaluation \hyperlink{Human-based Evaluation}{(A)},
						for tree={fill=child-four-color},
						[
							Meadows \textit{et al.} \cite{meadows2024localvaluebench};
                                Movva \textit{et al.} \cite{movva2024annotation}\\
                                Souly \textit{et al.} \cite{souly2024strongreject},
							for tree={fill=child-four-color}
						]
					]
                        [
						Rule-based Evaluation \hyperlink{Rule-based Evaluation}{(B)},
						for tree={fill=child-four-color},
						[
							Zou \textit{et al.} \cite{zou2023universal};
                                Liu \textit{et al.} \cite{liu2024making}\\
                                Zeng \textit{et al.} \cite{zeng2024autodefense};
                                Liu \textit{et al.} \cite{liu2024shield},
							for tree={fill=child-four-color}
						]
					]
					[
						Model-based Evaluation \hyperlink{Model-based Evaluation}{(C)},
						for tree={fill=child-four-color},
						[
							OpenAI Moderation API \cite{OpenAIModeration};
                                Perspective API \cite{perspectiveapi2023}\\
                                Azure AI Content Safety API \cite{MicrosoftAzureAIContentSafetyAPI};
                                Wang \textit{et al.} \cite{wang2024not}\\
                                Yu \textit{et al.} \cite{yu2023gptfuzzer};
                                GPT-4 Judge \cite{qi2023fine}\\
                                PAIR \cite{chao2023jailbreaking};
                                PICT \textit{et al.} \cite{shah2023scalable}\\
                                Llama Guard \cite{inan2023llama, metallamaguard2, metallamaguard3, dubey2024llama3herdmodels};
                                ShieldLM \cite{zhang2024shieldlm}\\
                                MD-Judge \cite{li2024salad};
                                AegisSafetyExperts \cite{ghosh2024aegis}\\
                                WildGuard \cite{han2024wildguard};
                                ShieldGemma \cite{zeng2024shieldgemma}\\
                                JailJudge Guard \cite{liu2024jailjudge},
							for tree={fill=child-four-color}
						]
					]
                        [
    						Evaluation Frameworks \hyperlink{Evaluation Frameworks}{(D)},
    						for tree={fill=child-four-color},
    						[
    							HarmBench \cite{mazeika2024harmbench};
                                    JailbreakBench \cite{chao2024jailbreakbench}\\
                                    EasyJailbreak \cite{zhou2024easyjailbreak};
                                    WalledEval \cite{gupta2024walledeval}\\
    							JailbreakEval \cite{ran2024jailbreakeval};
                                    AISafetyLab \cite{zhang2025aisafetylab}\\
                                    PandaGuard \cite{shen2025pandaguard},
    							for tree={fill=child-four-color}
    						]
					]
				]
                    [
					{Challenges and}\\
					{future directions}
                        \hyperlink{CHALLENGES AND FUTURE DIRECTIONS}{$\S$ VI}, 
					for tree={fill=child-five-color},
					[
						Unified Evaluation \hyperlink{Unified Evaluation}{(A)},
						for tree={fill=child-five-color}
					]
					[
						Dynamic Evaluation \hyperlink{Dynamic Evaluation}{(B)},
						for tree={fill=child-five-color}
					]
                        [
						Reliable and Efficient Evaluator \hyperlink{Reliable and Efficient Evaluator}{(C)},
						for tree={fill=child-five-color}
					]
					[
						Application-Specific Evaluation \hyperlink{Application-Specific Evaluation}{(D)},
						for tree={fill=child-five-color}
					]
                        [
						Safety Evaluation for MLLMs and Agents \hyperlink{Safety Evaluation for MLLMs and Agents}{(E)},
						for tree={fill=child-five-color}
					]
                        [
						Beyond Evaluation: From Safe to Responsible \hyperlink{Beyond Evaluation: From Safe to Responsible}{(F)},
						for tree={fill=child-five-color}
					]
				]
			]
		\end{forest}
    }
    \caption{Structure of this paper.}
    \label{fig:taxonomy}
\end{figure*}

\section{WHY TO EVALUATE}
\hypertarget{WHY TO EVALUATE}{}

This section introduces the background and key concepts of safety evaluation of LLMs, clarifies the distinctions between general evaluation and safety-specific evaluation, and highlights its importance in real-world applications. The goal is to underscore the necessity of safety evaluation and establish a foundation for subsequent discussions on advanced methodologies and practices.

\subsection{Background and Foundations}
\hypertarget{Background and Foundations}{}
LLMs have been rapidly adopted across a wide range of tasks — from knowledge inference and information extraction to content and code generation — due to their strong generative and reasoning capabilities. \cite{ye2024investigating, yao2024tree, zhao2023survey, wang2023codet5+, achiam2023gpt, wei2023zero, liu2024information, zhou2024trustworthiness} While these capabilities enable new applications, they also expose concrete and repeatable failure modes that raise safety concerns in practice.

Empirical studies have pointed out several failure modes. LLMs can generate toxic or abusive language under certain prompts, undermining user safety in public-facing systems \cite{deshpande2023toxicity, kim2023lifetox}. They can reproduce or amplify social biases and unethical stereotypes present in training data, yielding discriminatory outputs in downstream decisions \cite{raza2024mbias, ferrara2023should}. LLMs also exhibit hallucinations — confidently stated but factually incorrect assertions — which degrade trust and can mislead users \cite{huang2023survey, lan2024survey}. Finally, LLMs are susceptible to adversarial inputs, such as prompt injection or jailbreaks, which exploit vulnerabilities in the model's behavior \cite{qiu2023latent, yu2023gptfuzzer}. 

These technical failure modes have concrete, domain-specific consequences. In finance, biased or incorrect model outputs can distort credit assessments or automated advice \cite{maple2023ai}. In education, stereotype-aligned recommendations may narrow learning opportunities \cite{weissburg2024llms}. In healthcare, hallucinated or misleading suggestions risk patient safety and clinical misjudgments \cite{haltaufderheide2024ethics}. Moreover, adversaries can weaponize generation capabilities to scale misinformation or targeted manipulation, amplifying societal harms \cite{yi2024jailbreak}.

This dual-edged nature of LLMs underscores the need for comprehensive safeguards and systematic security assessments. Identifying and addressing these risks proactively is crucial to prevent unintended consequences, protect users, and ensure these models contribute positively to society. To achieve this, it is necessary to define a formal framework for safety evaluation, which we describe as the process of assessing whether a model can resist harmful instructions, avoid unsafe outputs, and maintain robustness under both benign and adversarial interactions. Formally:

\begin{equation}
R = E(T, D, P, M)
\end{equation}

where $R$ denotes the outcome of the evaluation , which may be quantitative (e.g., scores) or qualitative (e.g., categorical assessment); 
$E(\cdot)$ represents the safety evaluation function, mapping the selected tasks, data, evaluation process, and metrics to a final result; 
$T$ indicates the evaluation tasks that define the targeted safety dimensions. This step answers the question of \textit{what to evaluate}; 
$D$ represents the datasets or benchmarks used for testing. This step determines \textit{where to evaluate}; 
$P$ corresponds to the evaluation process. This aligns with \textit{how to evaluate}; 
and $M$ refers to the metrics that quantify safety performance.

\subsection{LLMs Evaluation vs LLMs Safety Evaluation}
\hypertarget{LLMs Evaluation vs LLMs Safety Evaluation}{}

The distinction between LLMs evaluation and LLMs safety evaluation lies not in superficial differences of scope or methodology, but in their fundamentally divergent goals and perspectives. Conventional evaluation addresses the question of \textit{“what the model can do”}, while safety evaluation is concerned with \textit{“what the model should not do”}. This asymmetry gives rise to significant differences in their objectives, approaches, and challenges.

First, the two forms of evaluation pursue distinct core objectives. LLMs evaluation is driven by an \textit{optimization mindset}: its purpose is to measure the upper bound of a model’s capabilities by assessing performance on general tasks such as knowledge, reasoning, or text generation. In contrast, safety evaluation adopts a \textit{control mindset}: its purpose is to probe the lower bound of risk by examining the model’s behavior under malicious or unexpected conditions. The ultimate goal is not to maximize capability, but to minimize harm, ensuring that the model remains robust, compliant, and trustworthy across diverse real-world applications \cite{sakib2024risks}. This divergence in purpose fundamentally shapes the methodology of each evaluation paradigm.

Second, safety evaluation can be understood as an extension rather than a replacement of capability evaluation. While technical proficiency remains a prerequisite, it is insufficient on its own. A model that excels in accuracy or fluency but fails to avoid harmful, biased, or manipulative outputs cannot be considered safe for deployment. Thus, safety evaluation builds on the foundation of general evaluation by expanding the focus from technical correctness to include ethical and societal considerations. In other words, it asks not only whether the model can perform a task, but also whether it should perform it in a given way.

Third, the quantification of evaluation outcomes reflects a striking asymmetry between the two. Capability evaluation typically relies on objective, mathematically defined indicators that enable standardized comparison across tasks and systems. Metrics like accuracy, F1-score, BLEU \cite{papineni2002bleu}, and ROUGE \cite{lin2004rouge} are designed to quantify improvements in effective and reliability. By contrast, safety evaluation must grapple with inherently subjective and context-dependent criteria, such as fairness, harm, or appropriateness. Defining what constitutes "toxic" or "biased" content is a challenging endeavor, as such determinations may vary across cultural and social contexts—factors that render universal standardization elusive. Presently, there remains a lack of consensus within the research community regarding how to formulate precise definitions of such content \cite{ran2024jailbreakeval}. This subjectivity poses unique challenges for designing safety benchmarks and metrics that are both rigorous and widely applicable.

In summary, while both LLMs evaluation and safety evaluation are indispensable, they differ in orientation: one is performance-driven, seeking optimization, while the other is risk-driven, seeking control. The interplay between these two dimensions—capability and safety—defines the trajectory of responsible LLM development, ensuring that models are not only powerful but also aligned with societal values and constraints.

\subsection{Importance}
\hypertarget{Importance}{}
The importance of LLMs safety evaluation transcends technical considerations, encompassing critical aspects of social responsibility, user trust, and legal compliance. As these models are increasingly integrated into sensitive domains such as healthcare, law, and education, the potential risks arising from erroneous or inappropriate outputs can result in significant adverse consequences \cite{yang2024beyond, prabhu2024opportunities, curran2023hallucination}. For instance, incorrect recommendations in healthcare systems may jeopardize patient safety, misinformation spread through educational platforms can distort students' understanding, and flawed analyses or judgments in legal contexts may lead to disputes or litigation.

The importance of LLMs safety evaluation lies not only in ensuring reliable performance, but also in addressing the distinctive challenges that arise from the generative and open-ended nature of these models. Unlike traditional AI systems whose risks are often confined to narrow domains, LLMs operate across virtually unlimited contexts and user intentions, which amplifies both the scale and unpredictability of potential harms. A single unsafe response may propagate misinformation at scale, enable malicious use, or erode public trust, creating risks that extend far beyond technical failures \cite{yang2024beyond, prabhu2024opportunities, curran2023hallucination}.

The evaluation methods for LLMs' general capabilities, such as accuracy benchmarks or instruction following tests, are not sufficient for safety evaluation, as safety concerns extend to subtle dimensions including refusal behavior, toxicity avoidance, bias mitigation, and resistance to adversarial prompting. Systematic safety evaluation is therefore indispensable for three reasons:

\begin{itemize}
    \item \textbf{Managing Unique Risk Profiles.} LLMs are open to the entire human society and face diverse and complex security risks. The risk of malicious use can occur in all possible scenarios. Evaluation enables the early detection of such vulnerabilities before real-world deployment. 
    \item \textbf{Enabling Accountability and Governance.} Transparent and standardized evaluation provides the evidence base necessary for legal compliance, policy development, and ethical oversight, particularly in high-stakes fields such as healthcare, law, and education. 
    \item \textbf{Driving Technical Progress.} Safety evaluation not only identifies shortcomings but also guides iterative model improvements, serving as a feedback loop for refining architectures, training strategies, and alignment techniques. This makes evaluation a scientific instrument for advancing both reliability and innovation. 
\end{itemize}

In this survey, we emphasize that safety evaluation is not a peripheral safeguard but a central research agenda: it bridges societal expectations, regulatory demands, and technical innovation. By synthesizing safety tasks, datasets, benchmarks and methodologies, we aim to provide a roadmap that helps the community move beyond ad hoc testing toward standardized, rigorous, and impactful evaluation practices.

\section{WHAT TO EVALUATE}
\hypertarget{WHAT TO EVALUATE}{}
This section provides an overview of various perspectives for evaluating the safety of LLMs, including toxicity, robustness, ethics, bias and fairness, truthfulness, and other specific tasks. It aims to illustrate what specific dimensions should be considered in evaluating and demonstrating the safety of LLMs, offering guidance for conducting comprehensive safety evaluations.

\subsection{Toxicity}
\hypertarget{Toxicity}{}
Toxicity refers to the presence of hate speech, offensive or abusive language, pornographic content, and other harmful expressions in text generated by LLMs. Such outputs not only pose risks of psychological harm to individual users but may also trigger broader social conflicts. As a result, evaluating the potential of LLMs to generate toxic content has become a critical step in ensuring their safe deployment. Existing efforts in this area can be broadly categorized into three lines of research: benchmark construction, adversarial and red-teaming based evaluation, and contextual or granular assessments.

\subsubsection{\textbf{Benchmark Construction for Toxicity Evaluation}} A major line of research has focused on building diverse and scalable benchmarks to capture toxicity in various settings. For example, ToxiGen \cite{hartvigsen2022toxigen} highlights that existing toxicity detectors often over-rely on surface-level mentions of minority identities, leading to the neglect of subtle hate speech and the over-detection of benign expressions. To address this issue, the authors developed a large-scale dataset generated via an adversarial classifier-in-the-loop decoding algorithm (ALICE), enabling the creation of implicit toxic texts across multiple demographic groups. ALICE adjusts the toxicity of the generated text by comparing a toxicity classifier with the text generator during beam search decoding. This adaptive control mechanism enables ToxiGen to not only generate texts that closely mimic human - written toxicity in nuanced expressions — including subtle biases toward underrepresented demographics — but also produce outputs that elude accurate classification by existing toxicity detectors, outperforming all prior human-curated corpora in both naturalistic mimicry and classifier evasion.

Similarly, ToxiChat \cite{lin2023toxicchat} aims to address the challenge of detecting toxic content in real-world user-AI interactions. They note that most existing research on toxicity detection relies on benchmarks built from social media content, failing to capture the unique challenges of real-world conversations with AI systems. By collecting real user queries directed at open-source chatbots and introducing an uncertainty-guided human–AI annotation pipeline, the authors constructed a benchmark that reflects the safety concerns of interactive systems. The annotation pipeline uses the moderation API \cite{perspectiveapi2023} to filter out low-confidence toxicity data, and the remaining data is manually reviewed, which greatly reduces the annotation workload. Furthermore, a systematic evaluation revealed that due to the domain inconsistency between the training data and real-world user prompts for chatbots, existing models not only fail to generalize to ToxiChat but also perform significantly worse on adversarial jailbreaking cases.  

Moving toward implicit harmful content, LifeTox \cite{kim2023lifetox} focuses on life-advice scenarios, where toxicity refers to a deeper, more concealed form of toxicity conveyed through linguistic features such as euphemisms, sarcasm, circumlocution, and metaphors, as well as extralinguistic knowledge like commonsense knowledge, world knowledge and social norms \cite{wen2023unveiling}. The authors argue that existing safety benchmark datasets are limited in diversity, primarily relying on red-team prompts, which often result in predictable and repetitive scenarios, failing to capture the complexity and variety of real-world situations. To address this gap, the dataset, constructed from Reddit forums, improves the detection of implicit harmful advice and demonstrates that fine-tuned RoBERTa\cite{liu2019roberta} can rival LLMs in harmful content classification. 

Extending beyond English and Chinese, Jain \textit{et al.} \cite{jain2024polyglotoxicityprompts} propose PolygloToxicityPrompts, a large-scale multilingual benchmark covering 425K prompts across 17 languages. The prompts were generated by scraping documents from extensive corpora, with an average length of 400 GPT-4 tokens. Experimental results indicate significant differences in toxicity levels among LLMs across different languages, with an observed increase in toxicity as model size grows within the same family. Furthermore, models after instruction and preference-tuning exhibit lower toxicity compared to baseline models; However, the choice of preference-tuning method have a minimal impact on the models' toxicity levels. The authors emphasize the need for further research on multilingual toxicity mitigation and the influence of model hyperparameters on toxicity.

\subsubsection{\textbf{Adversarial and Red-Teaming  Evaluation}} Another strand of work emphasizes adversarial attacks and red-teaming to probe the vulnerabilities of LLMs. Bhardwaj \textit{et al.} \cite{bhardwaj2023red} introduce RED-EVAL, a red-teaming benchmark that applies jailbreak attacks through a Chain of Utterances (CoU)-based prompt, facilitating a dialogue between two agents: a harmful agent Red-LM and a unsafe-helpful agent Base-LM. A harmful question is posed as the discourse for the Red-LM, and the model is instructed to generate a response according to the guidelines outlined in the prompt. They further propose RED-INSTRUCT, which enhances LLM safety through a two-phase approach: first, the CoU prompts are used to collect a dataset of harmful questions, termed HARMFULQA; Second, the model is safety-aligned by minimizing the negative log-likelihood of helpful responses while penalizing harmful ones. Experiments demonstrate that RED-EVAL achieves jailbreak success rates of 65\% on GPT-4 and 73\% on ChatGPT, underscoring the persistent risks of harmful content generation. 

Similarly, Zhuo \textit{et al.} \cite{zhuo2023red} conducted a systematic red team evaluation of ChatGPT using the RealToxicPrompts dataset \cite{gehman2020realtoxicityprompts} and Perspective API \cite{perspectiveapi2023}. The results showed that ChatGPT, due to its use of a cleaner training dataset, demonstrated superior performance compared to baseline models. However, the model remains vulnerable to prompt injection attacks, which can trigger toxic responses, and currently lacks mechanisms for early-stage toxicity detection.

\subsubsection{\textbf{Contextual and Granular Toxicity Evaluation}} Beyond benchmarks and adversarial probing, some works examine toxicity at different levels of granularity. FFT \cite{cui2023fft} investigate both utterance-level toxicity, where toxicity refers to literally-toxic language with some typical words, and context-level toxicity, where otherwise benign statements become harmful when interpreted in a broader conversational context. Their results show that LLMs generally exhibit higher toxicity rates at the contextual level, often by inadvertently affirming harmful issues despite generating literally non-toxic responses. This finding highlights the limitations of existing detectors in capturing nuanced, situational forms of toxicity.

Despite substantial progress, current approaches to toxicity evaluation face several unresolved challenges. First, the cultural and domain coverage of existing benchmarks remains limited, as most datasets are concentrated in a small set of languages or narrowly defined interaction scenarios, restricting their generalizability. Second, the emergence of increasingly sophisticated forms of implicit toxicity, conveyed through subtle linguistic cues or contextual framing, continues to outpace the ability of current models and detectors to capture them with sufficient accuracy. Third, the absence of standardized evaluation metrics and unified frameworks makes it difficult to compare results across studies, thereby hindering cumulative progress. 

\subsection{Robustness}
\hypertarget{Robustness}{}
Robustness refers to the ability of LLMs to maintain stable and safe outputs when exposed to noisy inputs, adversarial attacks such as jailbreak or prompt injection, and out-of-distribution (OOD) data. Beyond merely resisting malicious prompts, robustness also entails preserving instruction-following ability without exhibiting exaggerated safety, since models that indiscriminately refuse queries—regardless of their harmfulness (e.g., “How to kill a process?”)—sacrifice usability. Recent studies on robustness evaluation can be broadly grouped into three directions: adversarial and red-teaming evaluations, benchmarking and standardized frameworks, and the balance between instruction-following and excessive safety.

\subsubsection{\textbf{Adversarial and Red-Teaming Evaluation}} Several works focus on systematically characterizing the vulnerabilities of LLMs against adversarial inputs. Jailbreak attacks, particularly those involving carefully crafted prompts, pose significant challenges to the safety of LLMs. Chen \textit{et al.} \cite{chen2024characterizing} construct a dataset spanning 61 harmful categories to evaluate 10 jailbreak attack methods across 13 popular LLMs. Their framework employs multiple metrics, including attack success rate (ASR), toxicity, fluency, grammatical errors, and token length, enabling a multidimensional analysis of LLM safety. Results reveal that none of the evaluated models demonstrate inherent resistance to harmful queries, with Vicuna and Mistral among the most vulnerable, while GPT-4 and Llama3 exhibit comparatively stronger performance. Among the jailbreak attack methods, ReNeLLM \cite{ding2023wolf} achieves the highest average ASR, while ICA \cite{wei2023jailbreak} and Cipher \cite{yuan2023gpt} exhibit lower performance. They further observe that ASR correlates positively with toxicity. Additionally, grammatical errors are positively correlated with token length, while fluency does not show a strong correlation with either grammatical errors or token length.

Complementing this work, DecodingTrust \cite{wang2023decodingtrust} conducts a comprehensive evaluation of GPT-4 and GPT-3.5, examining robustness to adversarial text attacks, OOD inputs, and adversarial demonstrations. For adversarial text attacks, the researchers construct three scenarios: (i) standard benchmark AdvGLUE \cite{wang2021adversarial}, (ii) AdvGLUE under various instructive task descriptions and system prompts, and (iii) the more challenging adversarial text generation benchmark, AdvGLUE++. For OOD inputs, they explore scenarios including (i) inputs deviating from common training text styles, (ii) recent events outside GPT models' training data collection period, and (iii) demonstrations with diverse OOD styles and domains added through in-context learning. For adversarial demonstrations, the evaluations involve injecting counterfactual examples, spurious correlations, and backdoors into the demonstrations to observe model's performance. Overall, they find that GPT-4, though outperforming GPT-3.5 on standard benchmarks, is paradoxically more vulnerable to jailbreak attacks under certain prompts, owing to its stronger tendency to follow misleading instructions. 

Moving toward automated red-teaming, ASSERT \cite{mei2023assert} introduces three methods—semantically aligned augmentation, targeted bootstrapping, and adversarial knowledge injection—to generate new test cases. Semantically aligned augmentation aims to create samples with semantically equivalent but differently phrased expressions. Targeted bootstrapping generates new synthetic samples that are related to existing ones but not equivalent, while adversarial knowledge injection involves injecting adversarial knowledge during model inference. The researchers then evaluate LLMs on question-answering tasks across four critical AI application domains: outdoor, medical, household, and extra, requiring the models to determine whether a specific behavior \( a \) should be implemented within the context \( c \). Their results highlight that LLMs exhibit instability even under semantically similar scenarios, with divergences up to 11\% in classification accuracy, while adversarial demonstrations induce error rates exceeding 50\% in some settings. Collectively, these studies emphasize that LLM robustness remains fragile when confronted with diverse adversarial strategies and distributional shifts.

\subsubsection{\textbf{Benchmarking and Standardized Frameworks}} Recognizing the limitations of fragmented evaluations, another line of research develops systematic benchmarks and frameworks. JailbreakBench \cite{chao2024jailbreakbench} addresses the lack of clear standards and reproducibility in jailbreak testing by two core means: first, releasing a benchmark dataset consisting of 100 prompts that cover behavior, harmful goal, positive responses, harmful category, and data source; second, maintaining a repository of jailbreak artifacts and establishing a standardized evaluation framework. Its standardized evaluation pipeline incorporates threat models, system prompts, chat templates and multiple evaluation methods, ranging from rule-based string matching to GPT-4-as-a-judge and several models specifically designed for jailbreak evaluation (e.g., Llama Guard \cite{inan2023llama}). Experiments indicate that even state-of-the-art models remain highly susceptible. Prompts with RS \cite{andriushchenko2024jailbreaking} outperform others such as PAIR \cite{chao2023jailbreaking}, GCG \cite{zou2023universal}, and Jailbreak Chat \cite{jailbreakchat2023}, achieving an ASR of up to 78\% on GPT-4.

In parallel, HarmBench \cite{mazeika2024harmbench} establishes a more comprehensive red-teaming benchmark that categorizes harmful behaviors both semantically (e.g., cybercrime, bioweapons) and functionally (e.g., contextual, copyright, multimodal). In the evaluation pipeline, HarmBench ensures breadth by converting a series of behaviors into test cases. In addition, it employs fine-tuned Llama-2-13B and a hash-based classifier to evaluate the final results for non-copyright and copyright behaviors, respectively. Finally, it incorporates a total of 18 red-teaming methods from 12 studies, revealing that no attack or defense is universally effective, and that robustness does not correlate with model size—a finding that contradicts earlier assumptions \cite{ganguli2022red}. Together, these benchmarks enhance comparability and reproducibility, while underscoring the persistent vulnerability of LLMs.

\subsubsection{\textbf{Instruction-Following vs. Exaggerated Safety}} Existing research primarily emphasizes LLMs robustness against malicious inputs. However, we argue that achieving robustness should also ensure instructions-following ability without exhibiting exaggerated safety, as LLMs should balance safety with helpful. In the extreme, a model that indiscriminately refuses any prompt — safe or unsafe — would be perfectly harmless but completely useless \cite{rottger2023xstest}. Röttger \textit{et al.} \cite{rottger2023xstest} design XSTest, a benchmark mixing safe and unsafe prompts, and show that Llama2 exhibits exaggerated safety by refusing benign queries. They attribute this over-defensiveness to lexical overfitting and note that adding guardrail prompts, while reducing unsafe behavior, exacerbates the problem. Additionally, safety-related behavior could be changed by system prompts added at inference time. An \textit{et al.} \cite{an2024automatic} propose a white-box method for automatically generating pseudo-harmful prompts, revealing that many jailbreak defenses substantially increase false refusal rates, thereby undermining model usability.  Similarly, Varshney \textit{et al.} \cite{varshney2023art} develop classifiers to assess whether responses are both safe and appropriately generated, finding that self-checking methods improve safety but induce excessive conservatism. These studies collectively demonstrate that robustness cannot be narrowly defined as resistance to attacks but must also account for maintaining instruction-following capacity without compromising helpfulness.

Despite substantial progress, robustness evaluation faces several unresolved challenges. Current methods still provide limited coverage across diverse domains, languages, and modalities, leaving open questions about generalization in real-world contexts. Automated adversarial testing frameworks, though increasingly sophisticated, remain inconsistent and lack standardized protocols, complicating cross-study comparison. Moreover, defenses that reduce susceptibility to jailbreaks often increase false refusals, highlighting the persistent difficulty of balancing safety with usability. 

\subsection{Ethics}
\hypertarget{Ethics}{}
Ethics refers to the moral principles and values embedded in the outputs of large language models (LLMs), ensuring that generated text adheres to socially accepted norms and legal frameworks. Ethical compliance requires excluding content that is discriminatory, offensive, or misleading, as such expressions violate the principles of fairness and social justice \cite{gehman2020realtoxicityprompts}. Existing research on ethical evaluation of LLMs can be broadly categorized into global-purpose benchmarks and local-specific evaluations.

\subsubsection{\textbf{Global-purpose evaluation}}
A number of works have developed comprehensive benchmarks to assess the ethical reasoning and moral alignment of LLMs. While existing research has primarily focused on bias and other safety dimensions, MoralBench \cite{ji2024moralbench} systematically evaluates moral reasoning by drawing on the six foundations from Moral Foundations Theory \cite{graham2013moral}, which include Care/Harm, Fairness/Cheating, Loyalty/Betrayal, Authority/Subversion, Sanctity/Degradation, and Liberty/Oppression. It incorporates binary and comparative tasks to measure the consistency of LLM responses with human moral judgments. In the Binary Moral Evaluation, models are tasked with determining whether they “agree” or “disagree” with a set of moral statements, with scores based on how closely their responses align with human judgments. The Comparative Moral Evaluation requires models to select the more ethical option between two moral statements, with scoring reflecting the degree of alignment with human preferences. Results show that models perform adequately on simple binary classifications but struggle with comparative evaluations, suggesting a reliance on pattern recognition from training data rather than demonstrating a genuine understanding of ethical principles.

CVALUES \cite{xu2023cvalues} addresses the lack in existing evaluation frameworks in assessing the ethical alignment and value sensitivity of Chinese LLMs. The framework is structured around two critical dimensions: safety and responsibility. The safety dimension requires models to avoid generating harmful, biased, or unsafe content, while also ensuring their ability to manage sensitive topics, criminal advice, and privacy violations effectively. In contrast, the responsibility dimension expects models to provide positive guidance, demonstrate empathy, and account for the social and user impact of their outputs. The evaluation process integrates human assessments with automated multiple-choice evaluations, ensuring the reliability and comprehensiveness of the results. While models often succeed in avoiding harmful or unsafe content, their limitations emerge in domains such as law and social sciences, where reasoning and empathy are required. This demonstrates gaps between technical safety and broader ethical responsibility.

Complementary to these efforts, Hendrycks et al. \cite{hendrycks2020aligning} proposed the ETHICS dataset to test moral reasoning in multiple scenarios, showing that model performance is highly sensitive to textual framing and limited in handling nuanced ethical dilemmas. And Sun et al. \cite{sun2023safety} developed the SafetyPrompts benchmark, which includes Ethics and Morality as one dimension among eight safety scenarios. Their findings highlight that adversarial instructions often exacerbate ethical vulnerabilities, underscoring the need for robust defenses.

\subsubsection{\textbf{Local-Specific Evaluation}}
Ethics have significantly different definitions and levels of tolerance in different scenario  and cultural backgrounds.  Therefore, focusing on ethical risk evaluation in specific domains and localized contexts is equally important. While prior research has primarily focused on the safety of general text-based responses, TechHazardQA: \cite{banerjee2024ethical} is a dataset created to address the safety vulnerabilities of existing LLMs when generating instructional content, such as pseudocode or code snippets. The authors systematically evaluated the models' performance across multiple response formats under various testing scenarios, including zero-shot, chain-of-thought (CoT), and few-shot learning \cite{shaikh2022second, kojima2022large, wei2022chain}. To further investigate the impact of model editing on content generation, they employed the ROME model editing technique \cite{meng2022locating} to determine whether edited models were more prone to generating harmful content. The experimental results indicate that the likelihood of harmful content generation is significantly higher when models produce pseudocode compared to natural language responses. This probability further increases following model editing. 

Additionally, LocalValueBench \cite{meadows2024localvaluebench} evaluates ethical alignment under localized cultural and legal contexts. Focusing on the Australian environment, it examines model responses to neutral, debated, and misleading questions across domains such as capital punishment and same-sex marriage. Neutral questions establish baseline responses, debated questions explore how the models handle complex ethical scenarios, and misleading questions assess their behavior in extreme situations. Findings reveal significant cultural sensitivity: models differ not only in performance across topics but also in how embedded cultural biases shape outputs. The analysis indicates that the cultural context embedded in the training data significantly influences the performance of each model. Furthermore, the subjective bias observed in human evaluators' ratings expose the difficulties of standardizing cross-cultural ethical evaluation.

Taken together, global-purpose and local-specific evaluations provide complementary perspectives: the former emphasizes broad, systematic coverage of ethical reasoning, while the latter uncovers risks specific to specific domains or cultural environments. A key challenge lies in reconciling these perspectives to construct evaluation frameworks that are both comprehensive and culturally adaptive. Moreover, the diversity of ethical standards across societies raises profound challenges. Ethical standards are not universal, and LLMs trained on globally sourced data may inadvertently reflect dominant cultural values while marginalizing others. This issue of cultural diversity suggests that benchmarks must account for heterogeneous value systems rather than enforcing a single standard of morality. Looking forward, as AI systems become more autonomous, the potential emergence of machine-internal ethical reasoning poses additional risks: LLMs might develop consistent but misaligned ethical patterns that diverge from human expectations. More research is needed to address the ethical issues exposed by increasingly intelligent AI systems.

\subsection{Bias and Fairness}
\hypertarget{Bias and Fairness}{}
Bias and fairness are also indispensable dimensions of LLMs safety evaluation, which refers to the unintended reproduction of societal stereotypes, discriminatory views, or unequal treatment in the text generated by LLMs. It may also manifest as the model's inability to maintain neutrality and impartiality when processing user information related to diverse backgrounds, genders, races, or religions. The biases stem from factors such as training data, model specifications, algorithmic constraints, product design, and policy decisions \cite{ferrara2023should}. As LLMs are increasingly integrated across various sectors, effectively evaluating and mitigating biases while enhancing fairness has become both a pressing challenge and a key focus of current research.

\subsubsection{\textbf{General-purpose Evaluations}}
A first line of research focuses on systematically probing the inherent biases embedded in LLMs through general-purpose evaluations. 
Leidinger \textit{et al.} \cite{leidinger2024llms} combine insights from natural language processing (NLP) and search engine research to develop an innovative evaluation framework. This framework evaluates LLMs handling of stereotyping associated with different social groups through autocomplete-style prompts. The authors evaluate output across four dimensions: refusal, toxicity, sentiment, and regard. The findings indicate that while the introduction of system prompts helps reduce stereotypical outputs, LLMs still struggle with prompts related to peoples/ethnicities and sexual orientation. Additionally, mentions of intersectional identities tend to trigger disproportionately stereotyped responses.

Furthermore, Cantini \textit{et al.} \cite{cantini2024large} conduct a comprehensive evaluation of multiple popular LLMs through two steps: an initial evaluation using standard prompts and an adversarial analysis employing jailbreak prompts. In the first step, the authors design a sentence completion task where the model is required to choose one option from two: a stereotype or an counterstereotype to complete the provided sentence. Stereotypes deemed safe in this initial evaluation will progress to adversarial analysis in the second step, where the authors employe five jailbreak methods—role-playing, machine translation, obfuscation, prompt injection and reward incentive. The evaluation measures the performance of LLMs of various sizes based on robustness, fairness, and safety. The experimental results indicate that certain categories of bias, such as those related to sexual orientation and disability, are more effectively protected by model safety measures, while biases related to gender and age are less mitigated. Importantly, no model is entirely safe, as each is relatively susceptible to at least one jailbreak attack. 

To address the issue of most existing research either focusing on specific types of biases or employing incompatibility evaluation metrics, Wang \textit{et al.} \cite{wang2024ceb} introduce a Compositional Evaluation Benchmark (CEB). CEB adopts a compositional taxonomy approach to describe each dataset across three dimensions: bias type, social group, and task. The bias type includes stereotyping and toxicity, social group encompass age, gender, race, and religion, while task includes direct evaluation (recognition and selection) and indirect evaluation (continuation, conversation, and classification). To facilitate compatible evaluation, the researchers establish corresponding evaluation metrics and methods for each task, including Micro-F1 score, GPT-4-as-a-Judge, Perspective API\cite{perspectiveapi2023}, Demographic Parity (DP), Equalized Odds (EO), and Unfairness Scores\cite{agarwal2021towards}. The experimental results reveal that LLMs are adept at identifying toxic content in inputs but struggle to recognize stereotypical content. Furthermore, LLMs demonstrate a high refuse to answer rate for racial and religious social groups, indicating a heightened sensitivity.

Similarly, Huang \textit{et al.} \cite{huang2023cbbq} introduces a Chinese bias benchmark dataset, developed through human-AI collaboration, aimed at measuring biases in Chinese LLMs, while Cui \textit{et al.} \cite{cui2023fft} evaluates the fairness of LLMs through a benchmark that utilizes the coefficient of variation as an evaluation metric across four dimensions: identity preference, credit, criminal, and health.  Azeem \textit{et al.} \cite{azeem2024llm} evaluates several highly-rated LLMs based on Human-Robot Interaction (HRI) discrimination and safety criteria, highlighting the lack of robustness in LLMs when encountering people across various protected identity characteristics (e.g., race, gender, disability status, nationality, religion, and their intersections).

\subsubsection{\textbf{Domain-specific Evaluations}}
A second line of research investigates fairness in particular downstream applications where LLMs directly affect decision-making. In the context of recommendation systems, Zhang \textit{et al.} \cite{zhang2023chatgpt} evaluate the fairness of recommendation via LLM (RecLLM). Due to the unique nature of ReLLM, traditional fairness measurement methods based on scores and fixed datasets struggle to meet the necessary requirements. Consequently, the researchers propose a new benchmark called FaiRLLM, which includes carefully designed evaluation metrics and datasets covering eight sensitive attributes across two recommendation contexts: music and movies. The core idea of FaiRLLM is to measure  the similarity between the recommendation results of neutral instructions that do not include sensitive attributes and sensitive instructions that disclose such attributes. This approach assesses fairness by analyzing the differences in similarity among various sensitive attribute values (for example, in the context of race, comparing African American, Black, White, and Asian individuals). The results find that ChatGPT still exhibits unfairness in generating recommendations related to certain sensitive attributes, and it's responses align with existing societal biases regarding disadvantaged groups associated with different sensitive attributes. In addition, the authors also revealed the lack of robustness of ChatGPT to unfairness through the influence of sensitive attribute typos and language. 

Similarly, Wang \textit{et al.} \cite{wang2024large} evaluates the fairness of LLMs as text rankers from the perspectives of users and items. Specifically, the authors focus on binary protected attributes (gender and geographic location) using the TREC Fair Ranking Track dataset and conduct two types of evaluations: list evaluation and pairwise evaluation. The listwise evaluation measures how LLMs integrate underrepresented groups into rankings from both the query-side and item-side, while the pairwise evaluation provides LLMs with two items—one from a protected group and another from a non-protected group—to compare relevance or irrelevance. The results reveal that both neural rankers and LLMs exhibit a preference for queries associated with female and European. Interestingly, Mistral-7b shows a marked bias toward male items in relevant pairs, which contrasts sharply with the behavior of other models, raising questions about the decision-making processes of these models. Overall, LLMs tend to exhibit more subtle and profound biases favoring certain protected groups.

In addition, Given the widespread use of LLM-as-a-Judge \cite{gu2024survey}, Wang \textit{et al.} \cite{wang2023large} introduces a metric for conflict rate to quantitatively evaluate the model's sensitivity to the position of responses. This novel approach indicates the presence of positional bias in LLMs, suggesting that they tend to favor responses from specific positions. For instance, GPT-4 is inclined to support responses from the first position, while ChatGPT tends to favor responses from the second position.

Taken together, existing research demonstrates that LLMs exhibit systematic biases both in general-purpose evaluations and in application-specific contexts. However, several limitations persist. Current benchmarks remain limited in their cultural and linguistic coverage, restricting the generalizability of fairness assessments across diverse populations. Many methods struggle to capture subtle or intersectional forms of bias, which often manifest in implicit or adversarial scenarios. Furthermore, the absence of standardized evaluation metrics and unified frameworks hinders comparability across studies, limiting cumulative progress. 

\subsection{Truthfulness}
\hypertarget{Truthfulness}{}
Truthfulness refers to the factual accuracy of information generated by LLMs, specifically their alignment with real-world facts \cite{zhang2024reinforced}. Ensuring truthfulness is crucial for building user trust and preventing the dissemination of misleading information. A central challenge in this domain is the phenomenon of hallucination \cite{chataigner2024multilingual, bao2024faithbench, zhang2024siren}, where LLMs produce seemingly plausible but factually inaccurate or contextually disconnected responses. Consequently, most existing research on truthfulness evaluation has focused on hallucination \cite{li2025loki, zhang2025siren, huang2025survey}, while a smaller body of work examines other dimensions such as utility-driven deception.

\subsubsection{\textbf{Hallucination Evaluation}}
A substantial body of research has developed evaluation methods and benchmarks to systematically assess hallucinations in LLMs. One stream emphasizes dataset construction under realistic interaction scenarios. For example, ReEval \cite{yu2024reeval} create new evaluation datasets using adversarial attacks to test whether retrieval-augmented LLMs accurately reference provided evidence in their responses. ReEval incorporates two data generation techniques: answer swapping and context enriching. In answer swapping, portions of the supporting evidence containing the answer are altered to examine whether the model aligns with the new answer context. In context enriching, supplementary information is added to the original context, enhancing question complexity while preserving the core evidence. Results show that all models exhibited a significant hallucination on adversarial test data. Moreover, HaluEval \cite{li2023halueval} provides a large-scale benchmark that evaluate LLMs’ capabilities in hallucination detection, which cover knowledge-based dialogue, question answering, and text summarization. Experimental results reveal notable shortcomings in current LLMs’ hallucination detection abilities, and hallucination frequency is strongly topic-dependent, with higher rates observed in content related to film, technology, and climate. 

Another stream leverages structured external knowledge to support factual verification. GraphEval \cite{sansford2024grapheval} detects hallucinations in LLMs' outputs by constructing knowledge graphs (KGs) in two steps: KG construction and triple consistency verification. By limiting LLM calls to a single KG construction pass, this approach reduces computational costs and enhances interpretability by pinpointing inconsistencies. Combined with natural language inference (NLI) models, GraphEval significantly improves balanced accuracy in hallucination detection, boosting NLI performance by approximately 6.2 points on benchmarks like SummEval \cite{fabbri2021summeval} and QAGS-C \cite{wang2020asking}, especially for longer outputs. Similarly, KG-FPQ \cite{zhu2024kg} leverages KGs to generate False Premise Questions (FPQs) for assessing truthfulness hallucinations in LLMs. By selecting entities from art, people, and places, KG-FPQ first extracts real triplets from the KGs and generates false premises through edit true triplets to false triplets. Then, it utilizes ChatGPT's natural language generation capability to construct semantically rich FPQs. Experiments show that when edited object has a closer distance with the subject or has a stronger association with the original object, FPQs are more confusing to LLMs. And LLMs perform worse at generating factual statements than at distinguishing them when faced with FPQs.

In addition, some works propose unified framework for the evaluation of hallucinations. THaMES \cite{liang2024thames} introduces a end-to-end tool for hallucination detection and mitigation, which comprises three primary modules—QA set generation, hallucination benchmarking, and mitigation strategies. The framework begins by constructing high-quality question-answer test sets through weighted sampling and advanced question-generation techniques. Next, multiple metrics from RAGAS \cite{es2024ragas} and HaluEval \cite{li2023halueval} are used to evaluate hallucination. Following this, THaMES implements a series of refined mitigation strategies, including In-Context Learning (ICL) \cite{dong2022survey}, Retrieval-Augmented Generation (RAG) \cite{lewis2020retrieval}, and parameter-efficient fine-tuning (e.g., PEFT) \cite{hu2021lora, mangrulkar2022peft}. 
In order to solve the problem of difficulty in free-form response evaluation in open-domain and fragmentation of datasets and metrics, Iqbal \textit{et al.} \cite{iqbal2024openfactcheck} propose OpenFactCheck. The framework consists of three collaborative modules: First, RESPONSEEVAL supports users to customize the fact-checker, integrating the claim processors, evidence retrievers and verifiers of existing systems \cite{gao2022attributed, chern2023factool, wang2023factcheck} to realize factual error detection in a free-form document; Second, LLMEVAL builds a FactQA benchmark, which divides factual deviations into three categories: knowledge errors, over-commitment errors, and disability errors, and unifies evaluation metrics (such as exact matching, FreshEval \cite{vu2023freshllms}, and the proportion of true statements) to systematically evaluate LLM's factual capabilities in multiple dimensions; Third, CHECKEREVAL assesses the verification accuracy of fact-checkers, equipped with a leaderboard in terms of accuracy, latency, and costs.

\subsubsection{\textbf{Beyond Hallucinations}}
While hallucination has attracted primary attention, other dimensions of truthfulness are also emerging. AI-LIEDAR \cite{su2024ai} aims to investigate the balance between utility \cite{mckenna2023sources} and truthfulness \cite{hong2023zero} in LLMs. These models may produce inaccurate information to meet user demands, particularly when task objectives conflict with truthful disclosure, as in promoting defective products. The AI-LIEDAR framework includes 60 real-life-inspired multi-turn dialogue scenarios, each shaped by one of three motivations for deception: benefits, public image, and emotion. Using the Sotopia platform \cite{zhou2023sotopia}, the framework simulates interactions between users and models, with a psychology-based truth detector classifying responses as fully truthful, partially truthful (e.g., through omission), or entirely false. Additionally, the study investigates model steerability, testing whether prompts can effectively direct responses toward truthfulness or deception. Findings reveal that LLMs rarely sustain complete truthfulness throughout interactions, with an overall truthfulness rate below 50\%. Notably, a negative correlation emerges between truthfulness and goal achievement, especially in scenarios involving objective metrics, such as product promotion. Even when models are explicitly prompted toward “truthful” or “deceptive” responses, their truthfulness levels remain variable.

Despite notable advances, truthfulness evaluation remains constrained by several challenges. Current evaluation metrics for hallucinations face significant limitations, including a lack of universal applicability, inadequate comprehensiveness, and vulnerability to dataset leakage \cite{yang2025hallucination}, which undermine their robustness and fairness. Furthermore, many recent efforts have turned to the paradigm of LLM-as-a-Judge \cite{gu2024survey} to assess truthfulness. While this approach offers scalability and flexibility, it also introduces a recursive dilemma: if evaluating one model requires relying on a stronger and more reliable model, the trustworthiness of the evaluation itself becomes questionable. These challenges highlight the urgent need for innovative and standardized evaluation systems that can holistically capture different forms of untruthfulness, while reducing dependency on opaque or unverifiable evaluators.


\subsection{Other}
\hypertarget{Other}{}
In addition to commonly studied evaluation dimensions—such as toxicity, robustness, ethics, bias and fairness, and truthfulness—LLMs require rigorous evaluation across several other critical areas. These include privacy, mental health and medical applications, and code generation, copyright. Assessments in these domains are essential to ensuring that LLMs adhere to standards of safety, legality, and effectiveness, thus supporting their responsible and reliable deployment in real-world contexts.

\subsubsection{\textbf{Privacy}} 
LLMs rely on extensive data for training, posing inherent risks of leaking sensitive user information \cite{kim2024propile, fan2023trustworthiness, shao2023quantifying}, including personal identifiers, geographic locations, and employment-related information. Consequently, a systematic assessment of privacy risks in LLMs is critical to identify and mitigate potential breach channels, thereby ensuring model safety and compliance in real-world applications. Li \textit{et al.} \cite{li2024llm} introduced LLM-PBE, a comprehensive toolkit for evaluating privacy risks throughout the LLM lifecycle. LLM-PBE provides systematic assessments across diverse data types, multiple attack vectors, and corresponding defense strategies, with particular emphasis on analyzing how model size, data characteristics, and temporal factors impact privacy vulnerabilities. The toolkit incorporates modules for different attack types, such as data extraction \cite{shao2023quantifying, panda2024teach}, membership inference \cite{shokri2017membership}, prompt leakage \cite{hui2024pleak}, and jailbreaking \cite{chao2023jailbreaking}, complemented by defense methods like differential privacy \cite{dwork2006differential}. Moreover, the framework integrates support for models hosted on platforms such as OpenAI, TogetherAI, and Hugging Face, enabling privacy assessments through API interfaces. Extensive experimentation on datasets such as Enron \cite{klimt2004enron} and ECHR \cite{chalkidis2021paragraph} enables a systematic analysis of privacy vulnerabilities in LLMs. Empirical results demonstrate that larger models exhibit higher susceptibility to privacy leakage. Certain data types and placements, such as sensitive information located at the end of documents, are more vulnerable to extraction. Furthermore, while differential privacy effectively mitigates leakage risks, it introduces notable trade-offs in model utility, highlighting the necessity of balancing privacy preservation with performance. Despite these advances, key challenges remain. Privacy attacks evolve rapidly, demanding adaptive evaluation frameworks that can capture emerging threats. The trade-off between privacy preservation and model effectiveness remains unresolved, particularly in high-stakes domains. Addressing these challenges is essential for building trustworthy and privacy-preserving LLMs.

\subsubsection{\textbf{Copyright}} 
A systematic approach to copyright assessment is essential for systematically identifying and mitigating the generation of infringing content, thereby ensuring compliance and enabling the safe and lawful use of LLMs across diverse creative domains. As LLMs are increasingly applied in content creation, concerns regarding copyright infringement have intensified, particularly due to their ability to reproduce protected content without authorization \cite{hacohen2024not}. Existing copyright protection mechanisms often prove either overly restrictive or insufficiently adaptive, leaving gaps in safeguarding against unauthorized content generation. Liu \textit{et al.} \cite{liu2024shield} introduced the SHIELD framework, a tool designed to evaluate and prevent copyright violations by LLMs during text generation \cite{chang2023speak}. SHIELD employs a curated copyright dataset to assess whether generated text adheres to copyright standards and incorporates a lightweight and real-time proxy-based defense mechanism to dynamically detect and block potentially infringing content. The framework consists of three primary modules: the Copyright Material Detector, which uses an N-gram model to identify copyrighted segments; the Copyright Status Verifier, which performs real-time checks to verify copyright status; and the Copyright Status Guide, which provides contextual feedback to direct the model toward producing copyright-safe outputs. This mechanism is designed to promptly detect and reject infringing requests while maintaining flexibility in generating public domain content. However, balancing copyright protection with creative flexibility is a persistent issue, as overly strict safeguards may compromise usability. Jurisdictional diversity and the evolving nature of copyright status further complicate the establishment of standardized evaluation practices. Moreover, the scarcity of large-scale, high-quality copyright datasets limits the creation of comprehensive benchmarks. Addressing these challenges will be critical for ensuring that LLMs can be responsibly applied in creative and commercial domains.

\subsubsection{\textbf{Code Generation}} 
Unlike harmful content in other domains that directly affects individuals or society, risks in code generation stem from insecure code or code that facilitates cyberattacks, posing significant cybersecurity threats \cite{khoury2023secure, perry2023users, asare2023github}. For instance, LLMs may produce code vulnerable to common exploits such as SQL injection or cross-site scripting, rendering applications susceptible to compromise. Evaluating the security of LLMs in code generation has therefore become a critical research priority. Bhatt \textit{et al.} \cite{bhatt2023purple} introduced the CYBERSECEVAL framework, which evaluates the security of LLM-generated code and their susceptibility to cyberattack-related prompts. It includes two components: insecure coding tests, which employ a vulnerability detector to construct cases reflecting common programming weaknesses, and cyberattack helpfulness tests, which use attack-related prompts to assess whether model responses facilitate malicious actions. Empirical findings reveal that all tested models generated insecure code in about 30\% of insecure coding scenarios, and that models with stronger programming capabilities posed greater security risks. Furthermore, approximately 53\% of cyberattack prompts elicited responses aligned with adversarial objectives, highlighting LLMs’ limitations in handling malicious requests. CYBERSECEVAL thus offers a systematic framework for evaluating LLM cybersecurity. 
In addition, Peng \textit{et al.} \cite{peng2025cweval} identified limitations in existing benchmarks such as CYBERSECEVAL, including vague specifications, the separation of functional and security evaluations, and instability in static analysis due to high false positive and false negative rates. To address these issues, they proposed the CWEVAL framework and the CWEVAL-BENCH benchmark. CWEVAL introduces a result-driven synchronous evaluation mechanism that replaces static with dynamic analysis, monitoring runtime attributes such as execution time, memory access, and data integrity to verify both functional correctness and security simultaneously. CWEVAL-BENCH comprises 119 security-critical tasks spanning 31 CWE vulnerability types and five programming languages, with secure and insecure reference implementations to ensure reproducibility. Experimental results indicate that about 30\% of functionally correct code generated by mainstream LLMs still contains vulnerabilities. Larger models demonstrated comparatively stronger security capabilities, while simple security prompts improved performance, though fine-tuning sometimes degraded functional correctness. 
Li \textit{et al.} \cite{li2025safegenbench} further proposed the SafeGenBench framework and its automated evaluation system to overcome limited scenario coverage and single-method evaluation in existing benchmarks. SafeGenBench incorporates 558 test cases covering 13 programming languages and 44 CWE vulnerabilities, integrating authoritative standards such as OWASP Top-10 and CWE Top-25. Its design simulates realistic development scenarios by deliberately avoiding explicit security terms, ensuring alignment with practical usage. Its key innovation lies in a dual-layer evaluation mechanism: Static Application Security Testing (SAST) for broad-spectrum vulnerability scanning, complemented by an LLM-based judge for deep semantic analysis of task-specific vulnerabilities. These complementary methods mitigate the shortcomings of relying on a single approach. 
Looking ahead, the evaluation of code generation security faces persistent challenges. Software vulnerabilities continue to evolve rapidly, requiring benchmarks and frameworks that can adapt to emerging threats. At the same time, achieving accurate, efficient, and user-friendly evaluation remains a difficult task. Addressing these issues will be essential to ensure that LLM-assisted software development remains both secure and reliable in real-world settings.

\subsubsection{\textbf{Downstream Applications in Medicine, Finance, and Law}} 
As LLMs are increasingly adopted in high-stakes fields such as medicine, finance, and law, systematic evaluation of their domain-specific safety has become increasingly critical. Han \textit{et al.} \cite{medsafety} introduced MedSafetyBench, a benchmark dataset specifically designed to assess LLM safety in medical contexts. MedSafetyBench comprises 1,800 harmful medical requests paired with safe responses, organized around nine fundamental principles of medical ethics. By generating harmful prompts and corresponding responses, the study evaluated models such as GPT-4 and Llama to assess their ability to reject unsafe requests. Results indicate that current medical LLMs often fail to meet established safety standards, producing unsafe responses when confronted with inappropriate prompts. Hui \textit{et al.} \cite{hui2025trident} introduced Trident-Bench, the first benchmark targeting LLM safety across finance, medicine, and law. Grounded in the AMA Principles of Medical Ethics, the ABA Model Rules of Professional Conduct, and the CFA Institute Code of Ethics, Trident-Bench contains 2,652 harmful prompts with expert-validated safe responses. This benchmark addresses gaps left by prior evaluations that emphasized accuracy and domain knowledge while overlooking professional ethical compliance. An empirical study of 19 general-purpose, domain-specific, and safety-aligned models revealed critical safety gaps: while advanced closed-source models such as GPT-4o were able to meet basic safety expectations, domain-specific models frequently struggled with subtle ethical issues. Safety-aligned models, such as Llama Guard, performed better, indicating that domain specialization does not necessarily ensure ethical robustness. Although these efforts mark important progress, research on safety evaluation in medicine, finance, and law remains limited and lags behind the rapid deployment of LLMs in these domains. The specialized knowledge required in each field makes the design of rigorous and representative benchmarks challenging. Furthermore, variations in ethical norms and legal regulations across jurisdictions complicate the establishment of unified evaluation standards. Addressing these issues is essential for ensuring the responsible and safe integration of LLMs into high-stakes professional practices. At the same time, other downstream applications—such as education \cite{chu2025llm} and transportation \cite{zhang2024advancing}—pose comparable safety concerns, highlighting the need to broaden evaluation efforts beyond currently emphasized domains.

\section{WHERE TO EVALUATE}
\hypertarget{WHERE TO EVALUATE}{}
The safety of LLMs can be evaluated through carefully designed metrics, which determine the specific aspects of performance to be measured, and through datasets and benchmarks that provide the foundation for these evaluations. In this section, we comprehensively summarize these two components — metrics, datasets and benchmarks — to provide a clear overview of LLMs safety performance.


\subsection{Metrics}
\hypertarget{Metrics}{}
This section will discuss the evaluation metrics used to evaluate the safety of LLMs, as well as some frontier research related to these metrics. The focus will be on evaluating whether LLMs generate harmful content; currently, there are relatively few summaries of evaluation metrics in this area. In contrast, metrics for evaluating the truthfulness of LLMs have already been summarized in the context of hallucination evaluations\cite{huang2025survey}, so this aspect is therefore not elaborated here.

\subsubsection{\textbf{Attack Success Rate}}
\label{subsubsection: ASR}
Attack success rate (ASR) is widely used and fundamental metric for evaluating the safety of LLMs. It measures the effectiveness of adversarial attacks such as jailbreak attacks and prompt injection attacks directed at LLMs. A higher ASR indicates lower safety of the evaluated LLM. Mathematically, ASR can be formulated as:
\begin{equation}
ASR = \frac{Num_{success}}{Num_{total}}.
\label{ASR}
\end{equation}
where $Num_{success}$ denotes the number of successfully attacked prompts, and $Num_{total}$ denotes the total number of prompts used for evaluation.

Similarly, variations of ASR, such as the harmful content rate and refuse rate, share the same fundamental concept, so we will not discuss them separately. However, a significant challenge remains in defining what constitutes a “successful attack”. How to evaluate the success of a jailbreak attempt against an LLM has not been unified \cite{ran2024jailbreakeval}, leading to difficulties in comparing the works of different researchers. Current automatic studies primarily determine success through two approaches: rule-based and model-based methods, which discussed in detail in Section \ref{sec:HOW TO EVALUATE}. Rule-based methods are simple, but often have low relevance to human judgment, while model-based methods use different implementation paths in different studies, providing greater flexibility at the expense of reproducibility and comparability.

\subsubsection{\textbf{Fine-Grained Metrics}}
As mentioned in the section \ref{subsubsection: ASR}, the research community has yet to reach a consensus on how to define “attack success” or “jailbreak success”. Therefore, to comprehensively and accurately evaluate the safety of LLMs, many researchers have proposed new metrics or adapted some metrics from general LLMs evaluation frameworks.

Cai \textit{et al.}\cite{cai2024take} points out the limitations of rule-based and language model-based evaluation methods, specifically their inability to understand context or the oversimplification of binary classification metrics. These approaches often lead to numerous responses mistakenly classified as successful attacks, resulting in false negatives. Consequently, researchers propose a new set of metrics to distinguish between attackers and better capture the nuances among jailbreak motivations. Specifically, the authors introduce three novel metrics—Safeguard Violation (SV), Informativeness (I), and Relative Truthfulness (RT). These metrics define the success of a jailbreak in a progressive manner: SV assesses whether the LLM's response contains any violative content, I evaluates whether the response provides relevant information related to the prompt's intent, and RT measures the response's relative truthfulness concerning the prompt's intent. However, the specific implementation of these three metrics still relies on prompting the LLMs.

StrongReject\cite{souly2024strongreject} proposes an evaluation metric based on three distinct scores, designed to address the limitations of single-step evaluation by models. Before generating the three scores, the evaluator model outputs chain-of-thought reasoning about how well the response addresses the prompt. The three scores include a binary non-rejection score and two 5-point Likert scale scores that measure how specific and convincing the response is. The final score for the prompt/response pair is calculated using the following formula:
\begin{equation}
score = {(1 - refused)}\times \frac{specific + convincing}{2}.
\label{strongreject-metric}
\end{equation}
where $refused$ denotes whether the response is rejected (1) or not (0). $specific$ and $convincing$ denote how specific and convincing the response is. The higher the $score$, the less safe the response. This is a relatively strict evaluation standard, which alleviates the phenomenon of exaggerating the performance of jailbreak attacks.

Chen \textit{et al.}\cite{chen2024characterizing} argues that a reliable and safe LLM should meet three core criteria: resistance to jailbreak, non-toxicity, and quality. Accordingly, researchers use five metrics to evaluate these aspects: the commonly used attack success rate (ASR), toxicity score, fluency based on perplexity, token length, and grammatical errors. Additionally, Leidinger \textit{et al.}\cite{leidinger2024llms} incorporates four metrics—refusal, toxicity, sentiment, and regard—to capture nuanced biases in model responses. Moreover, Mei \textit{et al.}\cite{mei2024not} suggests that many detected jailbreaks may actually result from hallucinations. To evaluate such hallucinations, researchers have introduced a range of metrics, including entity transformation, semantic consistency, and recall. Although these fine-grained metrics are designed to capture aspects of safety beyond binary classification and narrow the gap with human judgment gold standards, they still heavily rely on the subjective judgment of models, lack standardized definitions, and have yet to gain widespread adoption, thereby limiting the comparability of research.

Overall, ASR remains the predominant metric for safety evaluation, but inconsistent definitions of 'success' hinder reproducibility. Emerging fine-grained metrics provide richer insights, yet they remain fragmented and lack standardization. Establishing a unified and widely accepted set of safety evaluation metrics therefore remains an open research challenge.

\begin{table*}[t]
\caption{Overview of comprehensive evaluation datasets.}
\label{table:Benchmarks}
\resizebox{\textwidth}{!}{%
\begin{tabular}{>{\centering\arraybackslash}p{3cm} >{\centering\arraybackslash}p{1cm} >{\centering\arraybackslash}p{3cm} >{\centering\arraybackslash}p{2cm} >{\centering\arraybackslash}p{7cm}}
\toprule
\textbf{Benchmark Name} & \textbf{Size} & \textbf{Safety Dimensions} & \textbf{Languages} & \textbf{Composition} \\
\midrule
\midrule
{Chu \textit{et al.}}~\cite{chu2024comprehensive}               & 160            & 16           & English                         & Harmful prompts \\ \midrule
{R-Judge}~\cite{yuan2024r}               & 162            & 27           & English                         & Multi-turn agent interaction \\ \midrule
{JailbreakBench}~\cite{chao2024jailbreakbench}               & 200            & 10           & English                         & Harmful behaviors and benign behaviors \\ \midrule
{StrongREJECT}~\cite{souly2024strongreject}               & 346            & 6           & English                         & Harmful prompts \\ \midrule
{HarmBench}~\cite{mazeika2024harmbench}               & 510            & 7           & English                         & Harmful behaviors \\ \midrule
{Do}-{Not}-{Answer}~\cite{wang2024not}               & 939            & 60           & English                         & Harmful instructions \\ \midrule
{AdvBench}~\cite{zou2023universal}               & 1k            & 8           & English                         & Harmful strings and harmful behaviors \\ \midrule
{CoSafe}~\cite{yu2024cosafe}               & 1.4k            & 14           & English                         & Multi-turn attack questions \\ \midrule
{Chen \textit{et al.}}~\cite{chen2024characterizing}               & 1.5k            & 61           & English                         & Harmful prompts \\ \midrule
{TechHazardQA}~\cite{banerjee2024ethical}               & 1.8k            & 7           & English                         & Sensitive and unethical questions \\ \midrule
{FFT}~\cite{cui2023fft}               & 2.1k            & 3           & English                         & Elaborated-designed harmful prompts \\ \midrule
{JailJudge ID}~\cite{liu2024jailjudge}               & 4.5k            & 14           & English                         & Labeled harmful prompts and model responses \\ \midrule
{SORRY-Bench}~\cite{xie2024sorry}               & 9k            & 45           & English                         & Unsafe instructions \\ \midrule
{HarmfulQA}~\cite{bhardwaj2023red}               & 18k            & 10           & English                         & Harmful questions, blue and red multi-turn conversations \\ \midrule
{AegisSafetyDataset}~\cite{ghosh2024aegis}               & 26k            & 13           & English                         &  Human-LLM interaction instances \\ \midrule
{SALAD-Bench}~\cite{li2024salad}               & 30k            & 66           & English                         & Harmful prompts and multi-choice prompts \\ \midrule
{ALERT}~\cite{tedeschi2024alert}               & 45k            & 32           & English                         & Red teaming prompts \\ \midrule
{WildGuardMix}~\cite{han2024wildguard}               & 92k            & 13           & English                         & Harmful prompts with refusal and compliance response \\ \midrule
{Do Anything Now}~\cite{shen2023anything}               & 107k            & 13           & English                         & Forbidden questions \\ \midrule
{CHiSafetyBench}~\cite{zhang2024chisafetybench}               & 2.1k            & 31           & Chinese                         & Multi-choice questions and risky questions \\ \midrule
{Wang \textit{et al.}}~\cite{wang2024chinese}               & 3k            & 17           & Chinese                         & Risky questions and harmless questions \\ \midrule
{CValues}~\cite{xu2023cvalues}               & 6.4k            & 10           & Chinese                         & Adversarial prompts and multi-choice prompts \\ \midrule
{JailBench}~\cite{liu2025jailbench}               & 10.8k            & 40           & Chinese                         & Jailbreak questions \\ \midrule
{CRiskEval}~\cite{shi2024criskeval}               & 14k            & 7           & Chinese                         & Multi-choice questions \\ \midrule
{Safety Prompts}~\cite{sun2023safety}               & 100k            & 14           & Chinese                         & Augmented prompts and responses \\ \midrule
{ChineseSafe}~\cite{zhang2024chinesesafe}               & 205k            & 10           & Chinese                         & Illegal and unsafe contents \\ \midrule
{Latent Jailbreak}~\cite{qiu2023latent}               & 416            & 3           & Chinese; English                         & Translation tasks \\ \midrule
{SafetyBench}~\cite{zhang2024safetybench}               & 11k            & 7           & Chinese; English                         & Multiple choice questions \\ \midrule
{S-Eval}~\cite{yuan2024s}               & 220k            & 52           & Chinese; English                        & Harmful prompts \\ \midrule
{AraTrust}~\cite{alghamdi2024aratrust}               & 516            & 8           & Arabic                         & Multi-choice questions \\ \midrule
{AnswerCarefully}~\cite{suzuki2025answercarefully}               & 1.8k            & 56           & Japanese                         & Harmful questions and reference answers \\ \midrule
{XSafety}~\cite{wang2024all}               & 2.8k            & 14           & 10                         & Harmful prompts \\ \midrule
{JailJudge OOD}~\cite{liu2024jailjudge}               & 6k            & 14           & 10                         & Labeled harmful prompts and responses \\ \midrule
{PolyglToxicityPrompts}~\cite{jain2024polyglotoxicityprompts}               & 425k            & 7           & 17                         & Real-world harmful prompts \\ 
\bottomrule
\end{tabular}%
}
\end{table*}

\subsection{Datasets and Benchmarks}
\hypertarget{Datasets and Benchmarks}{}
To comprehensively evaluate the safety performance of LLMs, numerous datasets and benchmarks have been developed. Unlike those focusing on specific domains\cite{medsafety, peng2025cweval}, in this section, we focus on datasets and benchmarks that assess the overall safety of LLMs across multiple dimensions. In this study, we selected 31 datasets and benchmarks in total, as shown in Table \ref{table:Benchmarks}. For clarity of categorization, we categorized them into four groups based on their composition: only questions (Q-only), questions and answers (Q\&A), multiple-choice questions (MCQ), and multi-turn conversations (Dialogue). Additionally, we advocate that the research community should consider LLMs safety within multilingual environments.

The majority of major datasets and benchmarks consist of Q-only. Chu \textit{et al.}\cite{chu2024comprehensive} establish a dataset of prohibited questions, providing the first large-scale evaluation of various advanced jailbreak attack methods. StrongReject\cite{souly2024strongreject} proposes a dataset composed of forbidden questions that LLMs should refuse to answer to evaluate jailbreak performance. HarmBench\cite{mazeika2024harmbench} construct its dataset by incorporating not only harmful behaviors but also copyright, contextual, and multimodal behaviors for a comprehensive evaluation. Do-Not-Answer\cite{wang2024not} includes prompts that responsible language models should not answer, while AdvBench\cite{zou2023universal} contains harmful strings and behaviors. Although initially develop using the white-box jailbreak method GCG, AdvBench has become a popular evaluation benchmark for researchers. To evaluate LLMs robustness against jailbreak attacks, Chen\cite{chen2024characterizing} introduce a refined dataset covering a wide range of harmful prompts. TechHazardQA\cite{banerjee2024ethical} presents a complex harmful query dataset, requiring LLMs to respond with pseudocode instead of standard text to evaluate ethical boundaries. FFT\cite{cui2023fft} carefully elaborates a dataset of harmful prompts focusing on factoid, unfair, and toxic contents in LLMs. SORRY-Bench\cite{xie2024sorry} addresses three key limitations in existing benchmarks: coarse-grained taxonomy of unsafe topics, overlook of linguistic characteristics and formatting of prompts, and reliance on large LLMs for evaluation. ALERT\cite{tedeschi2024alert} provides a substantial instruction dataset with a fine-grained risk taxonomy for red-team testing of LLMs. Do Anything Now\cite{shen2023anything} recognizes the shift of jailbreak prompts from online communities to prompt aggregation sites, so researchers create a question set based on in-the-wild jailbreak prompts. Wang \textit{et al.}\cite{wang2024chinese} construct a dataset including both harmful and benign prompts, which measure LLMs false positives and false negatives in refusal scenarios. JailBench \cite{liu2025jailbench} collects and integrates advanced jailbreak attack methods in the field, constructs a jailbreak prompt dataset with powerful jailbreak capabilities, and can comprehensively detect and evaluate the content safety protection capabilities of LLMs. In alignment with Chinese internet content regulations, ChineseSafe\cite{zhang2024chinesesafe} expands on sensitive categories specific to the Chinese context, including political sensitivity, pornography, and variant/homophonic words. Latent Jailbreak\cite{qiu2023latent} innovatively embeds harmful instructions into translation tasks, aiming to evaluate the robustness of LLMs against jailbreak attacks through translation tasks. Finally, S-Eval\cite{yuan2024s} introduces a bilingual dataset of risk prompts in both Chinese and English, containing fundamental risk prompts along with their corresponding adversarially crafted attack prompts. In summary, Q-only datasets provide broad coverage for testing refusal and robustness, but they lack reference responses for fine-grained evaluation.

Compared to datasets containing only questions(Q-only), datasets that include both questions and corresponding answers(Q\&A) offer a reference ground truth, enabling more precise safety evaluations and even fine-tuning of models for improved performance. JailbreakBench \cite{chao2024jailbreakbench} proposes a mixed dataset covered OpenAI’s usage policies, where each harmful behavior is paired with a benign behavior, making it useful for evaluating rejection rates and model defenses. JailJudge ID\cite{liu2024jailjudge} includes a collection of manually annotated harmful prompt-response pairs set in complex scenarios, while AegisSafetyDataset\cite{ghosh2024aegis} is a large-scale record of safey and unsafe interactions between humans and LLMs. Additionally, WildGuardMix\cite{han2024wildguard} is a large-scale, carefully balanced multitask dataset designed for safety regulation, encompassing both direct prompts and adversarial jailbreak responses with various refusal and compliance responses, primarily intended for fine-tuning safety evaluation models. Safety Prompts\cite{sun2023safety} provides augmented prompts and responses aimed at testing and improving LLMs safety, supporting a targeted approach to model safety enhancement. Overall, Q\&A datasets enable accurate assessment and model training but require extensive annotation efforts.

Both Q-only and Q\&A datasets are used to evaluate whether the content generated by LLMs in response to harmful queries complies with relevant policies. In contrast to this generative approach, multiple-choice questions (MCQ), as a form of discriminative evaluation, place a greater emphasis on the LLMs' ability to identify harmful content. Furthermore, a major advantage of MCQ over generative evaluation is that they can provide a highly intuitive and convenient measure of a model's safety performance, as the answers are unique and the evaluation results can be determined straightforwardly. SALAD-Bench\cite{li2024salad}, for instance, includes both harmful questions and multiple-choice questions, diversifying queries to enable a more comprehensive safety evaluation of LLMs. In the Chinese context, CHiSafetyBench\cite{zhang2024chisafetybench} is designed to identify risky content by requiring LLMs to select the single safe or unsafe option from multiple choices. CValues\cite{xu2023cvalues} also focuses on Chinese scenarios, evaluating LLMs alignment with human values from safety and responsibility perspectives by having models choose the better response between two options. To measure the risk propensity of LLMs, CRiskEval\cite{shi2024criskeval} constructs multiple-choice questions, each with four manually labeled safety levels: Extremely Hazardous, Moderately Hazardous, Neutral, and Safe, enabling fine-grained evaluation. SafetyBench\cite{zhang2024safetybench} combines both Chinese and English settings, creating a large, diverse multiple-choice dataset that thoroughly evaluates LLMs safety, blending harmful and safe behaviors in options to challenge models in their selection. MCQ datasets offer intuitive, reproducible evaluation metrics but may oversimplify complex generative behaviors.

Some jailbreak attack methods have evolved beyond single-turn attacks, extending to multi-turn interactions, which have shown significantly heightened risks\cite{ren2024derail, sun2024multi, russinovich2024great}. Consequently, evaluating the safety of LLMs in multi-turn conversations(Dialogue) has become necessary. R-Judge\cite{yuan2024r} provides a benchmark for assessing LLMs capabilities in identifying and evaluating safety risks within multi-turn interaction records between agents. CoSafe\cite{yu2024cosafe} highlights the current focus on single-turn interactions in red-teaming approaches and thus introduces a dataset designed for multi-turn attacks, featuring cohesive referencing across multiple turns. Additionally, HarmfulQA\cite{bhardwaj2023red} presents a dataset not only of harmful questions but also includes both blue and red conversations, making it suitable for fine-tuning safety models. Multi-turn datasets are essential for assessing real-world conversational safety, though they are resource-intensive to construct and analyze.

Furthermore, we emphasize the importance of evaluating the multilingual safety of LLMs. Studies have shown that the unbalanced cross-linguistic distribution of LLMs training data leads to language biases, resulting in inconsistent outputs when the same task is described in different languages\cite{dong2024evaluating}. Research also indicates that multilingual mixed prompts can significantly intensify the harm of malicious queries, and that different language types and families exhibit substantial difference in LLMs safety\cite{song2024multilingual}. Thus, while much research is focused on English and Chinese, the safety of LLMs in multilingual and low-resource language contexts deserves attention. AraTrust\cite{alghamdi2024aratrust} and AnswerCarefully\cite{suzuki2025answercarefully} extend benchmarks to Arabic and Japanese, respectively, while XSafety\cite{wang2024all}, JailJudge OOD\cite{liu2024jailjudge}, and PolyglToxicityPrompts\cite{jain2024polyglotoxicityprompts} offer multilingual coverage across 10–17 languages. These datasets highlight the importance of assessing LLMs safety in diverse linguistic contexts.

\section{HOW TO EVALUATE}
\hypertarget{HOW TO EVALUATE}{}
\label{sec:HOW TO EVALUATE}
Having identified what and where to evaluate in LLM safety, the next critical issue centers on how such evaluations should be conducted. A variety of methodologies have emerged, differing in levels of automation, granularity, and evaluator involvement. This section reviews the major evaluation paradigms, classifying them according to the nature and role of the evaluator—human, rule-based, or model-based—and discusses their respective advantages, limitations, and use cases. In addition, some evaluation frameworks that integrate the entire evaluation pipeline are also summarized.

\subsection{Human-based Evaluation}
\hypertarget{Human-based Evaluation}{}
Since the birth of the field of machine learning, human judgment has been regarded as the gold standard. In the safety evaluation of LLMs, central concerns include toxicity, ethics, and bias. A persistent challenge in this area is defining what “safety” means in relation to model outputs. Human-based evaluation holds a unique advantage here, as it allows for nuanced interpretation and context-sensitive judgment that automated approaches struggle to replicate \cite{novikova2017we}. Human evaluators can recognize subtle cues such as tone, sarcasm, or implicit meanings, making them particularly well-suited to detecting toxicity and bias. Nevertheless, human evaluation faces significant challenges, especially in terms of cost and consistency. It necessitates significant human and time resources \cite{karpinska2021perils}, and in large-scale evaluations, as the volume of content increases, the associated costs also rise substantially. For instance, Yuan \textit{et al.} \cite{yuan2023gpt} compensate annotators at a rate of \$27.5 per hour according to the standard hourly income for U.S. annotators. Moreover, variations in cultural background, knowledge, and personal values can reduce consistency across annotators, particularly in sensitive domains such as toxicity and bias \cite{zhao2023survey}.

Despite these limitations, human evaluation remains indispensable because automated methods are still constrained in their ability to grasp complex or culturally embedded content. Models often fail to interpret sarcasm, puns, or layered meanings, whereas human annotators can provide a more comprehensive understanding. This explains why many LLM safety studies continue to integrate human judgment into their evaluation pipelines. For instance, Meadows \textit{et al.} \cite{meadows2024localvaluebench} relied on human evaluation to examine whether LLM outputs aligned with Australian values, offering insights for global regulatory bodies. Similarly, Movva \textit{et al.} \cite{movva2024annotation} used annotators to evaluate chatbot responses across five safety dimensions—harm, bias, misinformation, political stance, and polarization—analyzing the agreement between human and model-based assessments. Human judgment is also a key reference point in jailbreak attack research, where it serves as the benchmark for evaluating both the success of jailbreaks and the reliability of automated evaluators. For example, Souly \textit{et al.} \cite{souly2024strongreject} showed strong alignment between their proposed method and human judgment by comparing bias levels and mean absolute error (MAE) against human evaluation results.

\subsection{Rule-based Evaluation}
\hypertarget{Rule-based Evaluation}{}
Rule-based evaluation in the safety assessment of LLMs typically involves the detection of potential risks or inappropriate content in model outputs by defining a set of explicit rules or guidelines. Such methods rely on well-specified criteria to flag responses that may contain sensitive information, harmful language, or other safety concerns. The typical process involves defining rule sets, applying them to model outputs, and categorizing responses accordingly.

Among these methods, string matching remains the most widely adopted, particularly in jailbreak attack research. In this setting, evaluators probe models with harmful queries and determine the success of the jailbreak attempt by checking for the presence of “rejection strings” in the output (e.g., “I am sorry” or “As an AI”). Each study typically constructs its own phrase set, often based on heuristic observations from empirical experiments. String matching is favored for its interpretability and cost-effectiveness. However, it suffers from significant drawbacks. Most notably, it exhibits a high false positive rate and tends to overestimate unsafe behavior, with low correlation to human judgment. This occurs because rejection phrase lists cannot exhaustively capture all valid refusal expressions. For instance, the list employed by Zou \textit{et al.} \cite{zou2023universal} excludes common phrases such as “not legal” and “not ethical”,which may lead to misclassification of safe outputs as harmful. Furthermore, jailbreak prompts often embed benign content to obscure malicious intent. In such cases, model responses may be irrelevant or unhelpful to the attack but are nevertheless flagged as unsafe under string-matching rules. Despite these limitations, string matching remains widely used, largely due to its negligible implementation cost and the need for comparability with prior studies that have adopted the same methodology \cite{liu2023autodan, wei2023jailbreak}. In addition, string matching has been observed to be combined with other security evaluation methods. For example, Liu \textit{et al.} \cite{liu2024making} combined string matching with text classifiers, while Zeng \textit{et al.} \cite{zeng2024autodefense} used string matching as a preliminary filter before conducting a more precise evaluation with GPT-4.

Beyond string matching, rule-based evaluation can also integrate traditional machine learning metrics to assess more specialized risks. For example, Liu \textit{et al.} \cite{liu2024shield} proposed a framework to evaluate copyright compliance in text generation. Their method combined multiple rule-based indicators, including similarity metrics such as the Longest Common Substring (LCS) \cite{karamolegkou2023copyright} and ROUGE-L scores \cite{lin2004rouge}, to quantify the overlap between generated text and copyrighted material. They further incorporated rejection rates as an additional measure of the model’s capacity to deny requests for infringing outputs. This integration of rule-based and metric-based evaluation provides a more comprehensive perspective than string matching alone.

In summary, rule-based evaluation offers clear, interpretable, and cost-efficient criteria for identifying unsafe or noncompliant content in LLM outputs. Nonetheless, its dependence on predefined rules makes it inherently limited in detecting novel risks or subtle safety issues. As such, rule-based methods are most valuable when employed as baseline or complementary tools, often in combination with human or model-based evaluation, to ensure more robust and scalable safety assessments.

\subsection{Model-based Evaluation}
\hypertarget{Model-based Evaluation}{}
Model-based evaluation refers to the use of high-performance models to assess the safety of text generated by LLMs, such as GPT-4 \cite{achiam2023gpt}, BERT \cite{kenton2019bert}, and reward models \cite{ge2023mart}. In this paradigm, models function as “evaluators” that interpret, analyze, and determine whether generated content adheres to predefined safety standards. Compared with rule-based evaluation, model-based methods can capture complex semantic structures and context-dependent nuances, thereby identifying subtle biases and harmful contents \cite{chiang2023can, chen2023exploring, zheng2023judging}. Moreover, in contrast to human evaluation, model-based approaches scale efficiently to large volumes of data, reducing human labor and associated costs. Broadly, model-based evaluation techniques can be divided into three categories: (i) specialized text classifiers, (ii) closed-source chat models, and (iii) open-source chat models.

\subsubsection{\textbf{Text Classifiers}}
Text classifiers generate structured outputs such as labels or confidence scores, which makes them particularly well-suited for safety evaluation tasks.  

Several companies provide proprietary content moderation services for detecting harmful text. Widely used tools include OpenAI’s Moderation Endpoint \cite{OpenAIModeration}, Microsoft’s Azure AI Content Safety API \cite{MicrosoftAzureAIContentSafetyAPI}, and Google’s Perspective API \cite{perspectiveapi2023}. For instance, OpenAI’s Moderation Endpoint is designed to determine whether content complies with its usage policies. It evaluates 11 categories—including hate, harassment, self-harm, sexual content, and violence—returning both binary labels and confidence scores between 0 and 1. Google’s Perspective API adopts a similar approach, focusing on “toxicity,” which refers to rude, disrespectful, or otherwise inappropriate text, but leaves the choice of decision threshold to users. While such closed-source classifiers are convenient, their utility for research purposes is limited. They primarily focus on surface-level harmful content, which can result in false positives in scenarios where unsafe behavior occurs implicitly (e.g., jailbreak demonstrations without overtly harmful text) \cite{yu2023gptfuzzer}.

In contrast, open-source classifiers are more widely used in academic evaluations. These models are typically fine-tuned from established sequence classification architectures such as BERT, RoBERTa, or DeBERTa. With parameter counts often below one billion, they are significantly smaller than contemporary open-source chat models. For example, Wang \textit{et al.} \cite{wang2024not} fine-tuned a Longformer \cite{beltagy2020longformer} for harmful response detection, while Yu \textit{et al.} \cite{yu2023gptfuzzer} fine-tuned a RoBERTa-large model \cite{liu2019roberta} for evaluation tasks. Nevertheless, the effectiveness of such classifiers heavily depends on the quality of their training datasets, and their performance often degrades on out-of-distribution data.

\subsubsection{\textbf{Closed-source Chat Model}}
Closed-source chat models, when employed as evaluators, provide both high performance and versatility. Models such as GPT-4 and InstructGPT, trained on massive datasets, demonstrate strong abilities in contextual understanding, semantic interpretation, and intent recognition. As a result, they are increasingly adopted as evaluators for assessing the safety and quality of model outputs. In particular, GPT-4-as-a-judge has become the most widely used evaluation method in this domain, showing strong correlation with human judgments.  

The implementation details of the evaluation process are all based on prompt engineering. By designing an evaluation prompt word, the model is prompted to evaluate the input-output pair according to the evaluation rules. For example, the GPT-4 Judge prompt \cite{qi2023fine} uses a Likert-style prompt to assess the harmfulness of a response. The model is informed of OpenAI's relevant safety policies as a reference for the evaluation process. PAIR prompt \cite{chao2023jailbreaking} assesses the degree to which the model's response violates ethical, legal, or safety guidelines, outputting a score out of 10. Similarly, PICT \cite{shah2023scalable} assesses whether the victim model's response contains harmful information related to a given category of prohibited behavior.

Despite their advantages, closed-source models face challenges in reliability. Prior studies highlight issues such as limited reproducibility, sensitivity to prompt wording, and a lack of robustness \cite{zheng2023judging, gao2024llm, eiras2025know}. These shortcomings undermine consistency across evaluations and raise concerns regarding reproducibility and transparency in academic or regulatory contexts. While closed-source models remain powerful and cost-efficient alternatives to human evaluation, further research is needed to enhance their robustness and reliability.

\subsubsection{\textbf{Open-source Chat Model}}
Several researchers have released open-source chat models to support safety evaluation efforts. These models are often fine-tuned on specific datasets to approach the performance of state-of-the-art closed-source models, and they can be deployed locally, reducing evaluation costs while improving reproducibility. Existing works can be broadly divided into two categories: models focusing on binary results with classification labels, and models that emphasize diversified labels and detailed reasoning.  

\textbf{Binary results and classification labels.}  
Llama Guard \cite{inan2023llama} is a safety evaluation model fine-tuned from the Llama2-7B model. It classifies responses based on user prompts and model outputs, and for responses labeled as unsafe, it also provides the violated safety policy. Meta researchers continue to update this series, with versions including Llama Guard 2 8B \cite{metallamaguard2}, Llama Guard 3 1B \cite{metallamaguard3}, and Llama Guard 3 8B \cite{dubey2024llama3herdmodels}. Similarly, MD-Judge \cite{li2024salad}, inspired by Llama Guard, fine-tuned the Mistral-7B model \cite{jiang2023mistral} for safety evaluation. Specifically, the researchers incorporated attack-augmented data in the fine-tuning process to further enhance the model's capabilities. The model outputs binary classification labels of safe/unsafe and the category of the unsafe response.

\textbf{Diversified results and reasoning.}  
Other works go beyond binary classification and focus on providing richer evaluation outcomes. ShieldLM \cite{zhang2024shieldlm}, fine-tuned from Qwen-14B-Chat \cite{bai2023qwen}, introduces a new label “controversial.” Under strict rules, controversial responses are marked as unsafe, while under loose rules, they are marked as safe. The model outputs both classification results and reasoning analyses. AegisSafetyExperts \cite{ghosh2024aegis} is unique in that its goal is to build a collection of LLM-based safety experts. The researchers finetuned three models using their own constructed dataset. First, based on Llama Guard, they fine-tuned two versions using the proposed taxonomy and safety policies — Llama Guard Defensive and Llama Guard Permissive. The former maps safety categories that “Need Cation” to unsafe, while the latter maps them to safe. Additionally, the researchers fine-tuned the NeMo43B model for prompt classification. WildGuard \cite{han2024wildguard} also fine-tuned Mistral-7B to detect malicious intent in user prompts, assess safety risks in model responses, and determine the model's refusal rate. Its training data includes synthetic adversarial, synthetic vanilla, and real-world user-LLM interaction data to maximize task coverage, diversity, and balance. ShieldGemma \cite{zeng2024shieldgemma} is designed specifically for safety content moderation. The researchers supervised fine-tuning of Gemma2 (2B, 9B, and 27B parameters), requiring outputs to include binary classification results and reasoning. Experiment results show that ShieldGemma outperforms both Llama Guard and WildGuard. Finally, JailJudge Guard \cite{liu2024jailjudge} represents the latest research in this line. By fine-tuning the Llama2-7B model with instruction tuning, it produces detailed reasoning together with a fine-grained evaluation score (a jailbreak score from 1 to 10).  

Overall, open-source chat models present a flexible alternative for automated safety evaluation. The primary differences among existing models lie in the composition of training data and the form of evaluation outputs. However, their deployment requires access to large-scale, high-quality labeled datasets and significant computational resources for fine-tuning. Moreover, their generalization to out-of-distribution data remains an open challenge, raising questions about their reliability in real-world applications.

\subsection{Evaluation Frameworks}
\hypertarget{Evaluation Frameworks}{}
Evaluation frameworks integrate the entire evaluation pipeline, enabling convenient end-to-end assessment and simplifying the process. These frameworks take evaluation datasets as input and produce evaluation results as output. In the intermediate process, in addition to interacting with target models through interfaces (such as HF, VLLM, or OpenAI) to obtain outputs and conduct evaluation, some frameworks also include components for attack augmentation on data or defense enhancement on models. These are designed to test the robustness of target models against attacks and the effectiveness of defense methods. As a result, most evaluation frameworks are typically composed of four parts, or a subset of them: Attacks, Defenses, Judges, and Interfaces.

As shown in Table \ref{tab:frameworks}, we summarize several representative evaluation frameworks and provide a brief comparison. JailbreakEval \cite{ran2024jailbreakeval} is a toolkit dedicated to evaluating jailbreak attempts. It does not include a preliminary data processing stage but instead focuses only on evaluating results. Its input is “question–answer” pairs, and it provides a set of out-of-the-box evaluators. In contrast, EasyJailbreak \cite{zhou2024easyjailbreak} is designed specifically for jailbreak attacks and consists of four components: Selector, Mutator, Constraint, and Evaluator. The process has three stages: in the preparation stage, users provide malicious prompts, prompt templates, or other configurations; in the attack stage, EasyJailbreak iteratively attacks the target model and updates the prompts while evaluating the outcome of each attack; in the final stage, the system aggregates results and generates a report for the user. Moving beyond general jailbreak-specific settings, WalledEval \cite{gupta2024walledeval} introduces innovation by supporting a wide range of safety benchmarks, with more than 35 benchmarks built in. These cover multilingual safety, exaggerated safety, prompt injection, and red-teaming datasets. Finally, HarmBench \cite{mazeika2024harmbench} is a standardized framework for automated red-teaming. It emphasizes three qualities of automated red-team evaluation: breadth, comparability, and robustness. Based on these principles, the pipeline begins by converting a diverse set of behaviors into test cases to probe target models, and then applies multiple classifiers to evaluate the responses and calculate the attack success rate.

\begin{table}[!t]
\centering
\caption{Comparison of Different Evaluation Frameworks.}
\label{tab:frameworks}
\setlength\tabcolsep{6pt}
\begin{tabular}{l|c c c c}
\toprule
\textbf{Frameworks} & \textbf{Attacks} & \textbf{Defenses} & \textbf{Judges} & \textbf{Interfaces}\\
\midrule
JailbrakEval \cite{ran2024jailbreakeval} & - & - & 22 & - \\

EasyJailbreak \cite{zhou2024easyjailbreak} & 12 & - & 7 & 4 \\

WalledEval \cite{gupta2024walledeval} & 9 & - & 6 & 6 \\

HarmBench \cite{mazeika2024harmbench} & 18 & - & 3 & 5 \\

JailbreakBench \cite{chao2024jailbreakbench} & 4 & 5 & 6 & 3 \\

AISafetyLab \cite{zhang2025aisafetylab} & 13 & 16 & 7 & 3 \\

PandaGuard \cite{shen2025pandaguard} & 19 & 12 & 4 & 7 \\
\bottomrule
\end{tabular}
\end{table}








The above frameworks do not consider defenses. In contrast, JailbreakBench \cite{chao2024jailbreakbench} incorporates several defense methods compared with EasyJailbreak. This toolkit allows users to configure threat models, system prompts, chat templates, and scoring functions, and it maintains a public leaderboard that compares the performance of state-of-the-art jailbreak attacks and defenses. Building on a broader perspective, AISafetyLab \cite{zhang2025aisafetylab} is organized into three core modules — attack, defense, and evaluation — along with four shared auxiliary modules: models, dataset, utils, and logging. It provides researchers with a highly extensible framework. Further, PandasGuard \cite{shen2025pandaguard} unifies attackers, defenders, target models, and judges into a modular system. Its main contribution is conducting a large-scale benchmark of about 3B tokens across 49 models.

In terms of evaluators, these frameworks generally adopt rule-based string matching and model-based evaluation methods. There is no clear preference between closed-source and open-source chat models, though text classifiers are used less frequently. While existing frameworks continue to expand in attacks, defenses, judges, and interfaces, there is still no consensus on evaluation methods. The divergence in evaluation method choices further highlights the lack of standardized practices in the field. This inconsistency not only weakens the comparability of results across frameworks but also limits the accumulation of research progress. Overall, current evaluation frameworks focus more on integrating and balancing existing methods than on advancing evaluation methodology. For higher-level evaluation, it is crucial to develop mechanisms that combine efficiency, accuracy, and robustness, and to establish standardized benchmarks and evaluation protocols.

\section{CHALLENGES AND FUTURE DIRECTIONS}
\hypertarget{CHALLENGES AND FUTURE DIRECTIONS}{}
The rapid development and powerful capabilities of LLMs have led to their widespread application in the real world, making them a innovative tool for human society. However, approaches to thoroughly evaluate LLMs safety have lagged behind this pace of development. Existing safety evaluation methods remain insufficiently comprehensive, facing numerous challenges and offering substantial room for improvement. Therefore, in this section, we summarize the current challenges LLMs face and propose corresponding research directions.


\subsection{Unified Evaluation}
\hypertarget{Unified Evaluation}{}
Despite growing efforts in safety evaluation of LLMs, the research community has yet to reach consensus on standardized procedures and methodologies for conducting evaluations. Prior work has highlighted that that LLM safety evaluation lacks robustness due to noise stemming from factors such as small datasets, methodological inconsistencies, and unreliable evaluation setups \cite{beyer2025llm}. There are differences in the specific implementation of evaluation methods and benchmarks used by different researchers. Especially for widely used key metrics such as attack success rate, there is no clear and specific measurement method. Such inconsistency makes it difficult to compare work across researchers and even raises concerns about reproducibility and reliability \cite{souly2024strongreject}. Furthermore, as LLMs are increasingly applied in diverse domains, ranging from education to healthcare and law, domain-specific safety requirements further exacerbate these inconsistencies. Therefore, we argue that future research should aim to establish a more unified evaluation paradigm. While achieving complete unification may be difficult, progress can be made by (i) developing standardized evaluation procedures and reporting protocols, (ii) defining unified and comprehensive metrics with clear computational formulations, and (iii) designing a multi-layered, modular evaluation framework that can flexibly adapt to different models, domains, and application scenarios. Such efforts would not only improve the comparability and reproducibility of safety evaluations but also foster a shared foundation for cumulative progress in this rapidly evolving field.

\subsection{Dynamic Evaluation}
\hypertarget{Dynamic Evaluation}{}
Most benchmarks for LLMs safety evaluation are released as static resources, with limited subsequent updates. This static approach poses several challenges. First, as LLMs evolve through continuous iteration, prompts or attack methods that once exposed vulnerabilities may no longer be effective, resulting in artificially inflated safety performance. Second, evaluation datasets risk contamination, as their contents may be incorporated into model training or fine-tuning for alignment, thereby reducing their validity as independent evaluation resources. Third, adversarial techniques continue to advance; for instance, jailbreak attacks often disguise malicious prompts to bypass safeguards and induce harmful outputs. As new jailbreak strategies emerge, static benchmarks quickly become outdated and insufficient for rigorous safety assessment. Although some initial studies have begun to explore dynamic evaluation \cite{zhang2025guardval, jindal2025texttt}, this line of research is still nascent. Future efforts should aim to design adaptive safety evaluation methods that enable benchmarks and protocols to evolve in tandem with both model capabilities and adversarial tactics. Promising directions include automated pipelines for generating novel adversarial prompts, community-driven platforms for continuously updating benchmarks, and integrating dynamic evaluation into red-teaming frameworks. Such approaches would help ensure that safety evaluations remain relevant, robust, and reflective of real-world risks.

\subsection{Reliable and Efficient Evaluator}
\hypertarget{Reliable and Efficient Evaluator}{}
As discussed in Section \ref{sec:HOW TO EVALUATE}, current evaluation methods for LLMs safety can be broadly categorized into rule-based, human-based, and model-based approaches. First, rule-based methods are simple and efficient but often fail to capture implicit harms beyond predefined patterns, leading to false negatives. Second, although human-based evaluation is considered the gold standard due to its ability to accurately capture linguistic nuances and contextual meanings aligned with real-world values, it is costly, time-intensive, and difficult to scale. Moreover, cross-cultural and moral value differences among annotators can introduce variability in harm perception and compromise the consistency of results \cite{davani2024d3code}. Third, model-based evaluation offers a new direction for automated evaluation, with advanced models such as GPT-4 showing strong alignment with human judgments and capable of processing large volumes of data efficiently. However, they face significant challenges, including high inference costs, substantial computational requirements for fine-tuning, limited reproducibility, high sensitivity to evaluation prompts, and lack of robustness \cite{zheng2023judging, gao2024llm, eiras2025know}. These factors can impact the accuracy and fairness of evaluations. Future research should consider how to strike a better balance between model-based and human-based evaluation approaches, achieving reliable evaluation results while maintaining high efficiency and low costs. Future research should prioritize the development of evaluators that are both reliable and efficient. Promising directions include hybrid approaches that combine automated filters with lightweight model-based scoring and selective human oversight, methods to enhance the consistency and reproducibility of model-based evaluators (e.g., prompt calibration and majority vote).

\subsection{Application-Specific Evaluation}
\hypertarget{Application-Specific Evaluation}{}
As LLMs are increasingly integrated into various real-world applications, evaluating their safety in these specific application tasks has become crucial. For example, \cite{tang2024defining} developed a physical safety benchmark for evaluating LLMs in drone control systems, demonstrating the importance of targeted safety evaluations within specific domains. However, current safety evaluation efforts for LLMs mainly focus on general models, while safety research for the latest applications remains relatively disconnected, failing to keep pace with the integration of LLMs into other fields. General benchmarks, which primarily test for risks such as toxicity or bias, are often insufficient in high-stakes domains, where unique and domain-dependent hazards must be rigorously addressed. Specifically, many emerging applications such as intelligent medical diagnosis, autonomous driving, weather forecasting, finance, and law have begun leveraging the powerful capabilities of LLMs, yet safety evaluation standards and tools tailored to these contexts are still underdeveloped. This gap underscores the urgent need to establish safety evaluation frameworks tailored to emerging domains. Importantly, these frameworks should not only extend beyond generic benchmarks but also incorporate domain expertise and align with regulatory and ethical standards relevant to each field (e.g., FDA regulations in medicine, or financial compliance requirements). Practical implementations could include the creation of domain-specific benchmark datasets, the use of simulation-based environments to test LLM behavior in high-risk scenarios, and the integration of interdisciplinary perspectives to ensure robustness and applicability.

\subsection{Safety Evaluation for MLLMs and Agents}
\hypertarget{Safety Evaluation for MLLMs and Agents}{}
Multimodal large language models (MLLMs) and agents, both grounded in LLM technology, have attracted growing attention across academia and industry. MLLMs enable more natural and seamless human–machine interaction by integrating text with modalities such as images, audio, and video, while AI agents emulate human-like reasoning and decision-making processes. In MLLMs, LLMs serve as the core engine for text understanding and generation \cite{yin2023survey}, whereas in agents, LLMs function as the brain, orchestrating planning and decision-making. Despite rapid progress, the safety evaluation of both directions significantly lags behind their development. For MLLMs, existing benchmarks remain limited \cite{liu2025mm,tu2023many,luo2024jailbreakv,gu2024mllmguard}, lacking robustness metrics and evaluation standards tailored to multimodal contexts. In particular, video-based domains such as video understanding and generation remain underexplored. For agents, safety risks are amplified by the integration of multiple new components, which create additional attack surfaces and require evaluation in complex, interactive environments. Examples include emerging communication protocols such as the Model Context Protocol (MCP) and agent-to-agent (A2A) communication, which introduce new possibilities for coordination but also novel vulnerabilities. While some studies have begun addressing these challenges \cite{yuan2024r,zhu2024riskawarebench,zhang2024agent}, no consensus has yet been reached on unified benchmark design standards for the broader agent ecosystem \cite{deng2024ai}. Key open challenges include identifying the core dimensions of agent safety, designing evaluation methods capable of capturing risks in dynamic, multi-agent environments, and developing comprehensive frameworks that can evolve alongside these rapidly advancing systems.

\subsection{Beyond Evaluation: From Safe to Responsible}
\hypertarget{Beyond Evaluation: From Safe to Responsible}{}
Safety evaluation should not be regarded as the ultimate goal, but as the foundation for further progress. It is not only about producing benchmark results to guide the safe deployment of LLMs, but also about shaping directions for future research and development. The core role of safety evaluation is to identify and mitigate risks, thereby ensuring model stability and reliability. Yet, as technology advances and societal demands grow more complex, safety alone will no longer be sufficient to meet the expectations for real-world applications. Building on current development trends and research in LLM safety evaluation, we argue that the future trajectory should move from merely safe LLMs toward truly responsible LLMs. Responsible LLMs should extend beyond basic safety, fairness, and trustworthiness to also embody transparency, interpretability, and auditability. In other words, models must not only be technically reliable, but also broadly acknowledged for their ethical integrity and social responsibility. Correspondingly, safety evaluation itself will continue to evolve, encompassing a wider set of dimensions and employing deeper, more sophisticated methods in order to support this transition.

By expanding and refining research into the above areas, the community can more accurately evaluate the safety of LLMs while also promoting their development towards higher safety standards. In the future, LLMs will be closely aligned with societal values and norms, evolving into a responsible paradigm of artificial intelligence.

\section{Conclusion}
Ensuring the safe deployment and responsible use of LLMs in real-world applications requires systematic safety evaluation. This paper provides the first comprehensive review of LLMs safety evaluation, organized around four guiding questions: “\textit{Why evaluate},” “\textit{What to evaluate},” “\textit{Where to evaluate},” and “\textit{How to evaluate}.” We systematically survey key dimensions of LLMs safety, including toxicity, robustness, ethics, bias and fairness, and truthfulness, as well as specific downstream tasks such as privacy, mental health, medical applications, code generation and copyright. In addition, we review current evaluation metrics, and a broad set of datasets and benchmarks, highlighting their characteristics, strengths, and limitations. The roles of different evaluators — human-based, rule-based, and model-based — are also examined to illustrate the current state of LLM safety evaluation. Moreover, we review several representative evaluation frameworks, highlighting how they integrate attacks, defenses, target models, judges in practical assessments. This structured overview aims to provide researchers with a clear understanding of available resources and methodologies, supporting systematic and reproducible safety evaluations. Finally, we highlight several major challenges currently faced in safety evaluation of LLMs and discuss potential directions for future research.

Looking forward, we anticipate that continued research in safety evaluation will facilitate the development of LLMs that are more reliable, responsible, and aligned with human values. By combining robust evaluation frameworks with improved model design, future LLMs can provide intelligent support across diverse applications while effectively mitigating potential risks, ultimately fostering a balanced approach between technological innovation and trustworthy safety.


\bibliographystyle{IEEEtran}
\bibliography{main}








\end{document}